\documentclass[10pt]{article} % For LaTeX2e
\usepackage[accepted]{tmlr}
\usepackage{amsmath,amsfonts,bm}

\def\eqref#1{equation~\ref{#1}}
\def\1{\bm{1}}

\def\mA{{\bm{A}}}
\def\mB{{\bm{B}}}

\def\mI{{\bm{I}}}

\def\mW{{\bm{W}}}

\DeclareMathAlphabet{\mathsfit}{\encodingdefault}{\sfdefault}{m}{sl}
\SetMathAlphabet{\mathsfit}{bold}{\encodingdefault}{\sfdefault}{bx}{n}

\usepackage{dsfont}
\usepackage{hyperref}
\usepackage{url}
\usepackage{float} % float 패키지 포함
\usepackage{mathrsfs}

\usepackage{xcolor}
\usepackage{soul}
\usepackage[utf8]{inputenc}
\usepackage{multirow}
\usepackage{booktabs}       % professional-quality tables

\usepackage{caption}
\usepackage{subcaption}

\usepackage{wrapfig}
\usepackage{multirow}
\usepackage{adjustbox}
\usepackage{colortbl}
\usepackage{tikz}
\usepackage{soul}

\usepackage{xspace}
\usepackage{color} % to change font color
\usepackage{algorithm}
\usepackage{algpseudocode,algorithmicx}
\usepackage{multicol}
\usepackage{tablefootnote}
\usepackage{makecell}
\usepackage{subcaption}
\usepackage{caption}
\usepackage[toc,page]{appendix}

\newcommand*\HalfHatchedCircle[1][1ex]{%
    \begin{tikzpicture}
        \draw (0,0) circle (#1);
        \clip (0,0) -- (0,-#1) arc[start angle=270, end angle=450, radius=#1] -- cycle;
        \foreach \x in {-1.4,-1.2,...,1.4} {
            \draw[rotate around={45:(0,0)}] (\x*#1,-2*#1) -- (\x*#1,2*#1);
        }
    \end{tikzpicture}%
}

\newcommand*\emptycirc[1][1ex]{\tikz\draw (0,0) circle (#1);} 

\newcommand*\halfcirc[1][1ex]{%
  \begin{tikzpicture}
  \draw[fill] (0,0)-- (90:#1) arc (90:270:#1) -- cycle ;
  \draw (0,0) circle (#1);
  \end{tikzpicture}}
\newcommand*\halfcircc[1][1ex]{%
\begin{tikzpicture}
\draw[fill] (0,0)-- (270:#1) arc (270:450:#1) -- cycle ;
\draw (0,0) circle (#1);
\end{tikzpicture}}
\newcommand*\fullcirc[1][1ex]{\tikz\fill (0,0) circle (#1);} 
\usepackage{tikz}
\usepackage{amssymb}% http://ctan.org/pkg/amssymb
\usepackage{amsmath}
\usepackage{pifont}% http://ctan.org/pkg/pifont
\newcommand{\cmark}{\ding{51}}%
\newcommand{\xmark}{\ding{55}}%

\usepackage{duckuments}

\newcommand{\ie}{\textit{i}.\textit{e}., }
\newcommand{\eg}{\textit{e}.\textit{g}., }
\newcommand{\viz}{\textit{viz}., }
\newcommand{\algrule}[1][.2pt]{\par\vskip.5\baselineskip\hrule height #1\par\vskip.5\baselineskip}

\title{Confidence-aware Denoised Fine-tuning of \\ Off-the-shelf Models for Certified Robustness}

\author{\name Suhyeok Jang* 
      \email jasuhe900@kaist.ac.kr \\ 
      \addr Korea Advanced Institute of Science \& Technology (KAIST)
      \AND
      \name Seojin Kim*
      \email osikjs@kaist.ac.kr \\
      \addr Korea Advanced Institute of Science \& Technology (KAIST)
      \AND
      \name Jinwoo Shin 
      \email jinwoos@kaist.ac.kr \\
      \addr Korea Advanced Institute of Science \& Technology (KAIST) 
      \AND
      \name Jongheon Jeong 
      \email jonghj@korea.ac.kr \\
      \addr Korea University \\\\
      * The authors contributed equally.}

\newcommand{\Algname}{Fine-Tuning with Confidence-Aware Denoised Image Selection\xspace}
\newcommand{\ALgname}{FT\text{-}CADIS\xspace}
\newcommand{\LowLoss}{Confidence-aware selective cross-entropy loss\xspace}
\newcommand{\HighLoss}{Confidence-aware masked adversarial loss\xspace}

\definecolor{pinegreen}{rgb}{0.0, 0.47, 0.44}
\definecolor{cornellred}{rgb}{0.7, 0.11, 0.11}
\definecolor{cadmiumgreen}{rgb}{0.0, 0.42, 0.24}
\definecolor{spirodiscoball}{rgb}{0.06, 0.75, 0.99}
\definecolor{Red7}{rgb}{0.941, 0.243, 0.243}
\definecolor{Eqpink}{RGB}{241,241,214}
\definecolor{Green7}{RGB}{55, 178, 77}
\definecolor{aliceblue}{rgb}{0.91, 0.94, 0.97}
\definecolor{darkblue}{rgb}{0.83, 0.89, 0.97}
\definecolor{SJViolet}{RGB}{105,100,171}
\definecolor{SJRed}{RGB}{237,109,107}
\definecolor{tablegreen}{rgb}{0.91, 0.94, 0.97}

\hypersetup{citecolor=MidnightBlue, linkcolor=BrickRed}
\hypersetup{
  linkcolor = cornellred,
  citecolor  = cadmiumgreen,
  colorlinks = true,
}

\def\month{11}  % Insert correct month for camera-ready version
\def\year{2024} % Insert correct year for camera-ready version
\def\openreview{\url{https://openreview.net/forum?id=99GovbuMcP}} % Insert correct link to OpenReview for camera-ready version

\begin{document}
    \maketitle

\begin{abstract}

The remarkable advances in deep learning have led to the emergence of many off-the-shelf classifiers, \eg large pre-trained models. However, since they are typically trained on clean data, they remain vulnerable to adversarial attacks. Despite this vulnerability, 
their superior performance and transferability make off-the-shelf classifiers still valuable in practice, demanding further work to provide adversarial robustness for them in a \emph{post-hoc} manner.
A recently proposed method, \emph{denoised smoothing}, leverages a denoiser model in front of the classifier to obtain \emph{provable robustness} without additional training. However, the denoiser often creates \emph{hallucination}, \emph{i.e.}, images that have lost the semantics of their originally assigned class, leading to a drop in robustness. Furthermore, its noise-and-denoise procedure introduces a significant distribution shift from the original distribution, causing the denoised smoothing framework to achieve sub-optimal robustness. In this paper, we introduce \emph{\Algname (\ALgname)}, a novel fine-tuning scheme to enhance the certified robustness of off-the-shelf classifiers.
\ALgname is inspired by the observation that the \emph{confidence} of off-the-shelf classifiers can effectively identify hallucinated images during denoised smoothing. Based on this, we develop a confidence-aware training objective to handle such hallucinated images and improve the stability of fine-tuning from denoised images. 
In this way, the classifier can be fine-tuned using only images that are beneficial for adversarial robustness.
We also find that such a fine-tuning can be done by merely updating a small fraction (\ie 1\%) of parameters of the classifier. Extensive experiments demonstrate that \ALgname has established the state-of-the-art certified robustness among denoised smoothing methods across all \(\ell_2\)-adversary radius in a variety of benchmarks, such as CIFAR-10 and ImageNet.

\end{abstract} 

\section{Introduction}
\label{sec:introduction}
Despite the recent advancements in modern deep neural networks in various computer vision tasks \citep{radford2021learning,rombach2022high,kirillov2023segment}, they still suffer from the presence of \emph{adversarial examples} \citep{szegedy2013intriguing} \emph{i.e.}, a non-recognizable perturbation (for humans) of an image often fools the image classifiers to flip the output class \citep{goodfellow2014explaining}. Such adversarial examples can be artificially crafted with malicious intent, \emph{i.e.}, \emph{adversarial attacks}, which pose a significant threat to the practical deployment of deep neural networks. To alleviate this issue, various approaches have been proposed to develop \emph{robust} neural networks, such as adversarial training \citep{madry2018towards, wang2019improving} and certified defenses \citep{wong2018provable, cohen2019certified,li2023sok}.

Among these efforts, \emph{randomized smoothing} \citep{lecuyer2019certified, cohen2019certified} has gained much attention as a framework to build robust classifiers. This is due to its superior provable guarantee of the non-existence of adversarial examples, \emph{i.e.}, certified robustness \citep{wong2018provable, xiao2018training}, under any perturbations confined in a \(\ell_2\)-norm. Specifically, it builds a \emph{smoothed classifier} through taking a majority vote from a \emph{base classifier}, \emph{e.g.}, a neural network, under Gaussian perturbations of the given input image. However, it has been practically challenging to scale the model due to a critical drawback: the base classifier should be specifically trained on noise-augmented data \citep{lecuyer2019certified, cohen2019certified}.

Recently, \citet{lee2021provable,carlinicertified} have introduced \emph{denoised smoothing} which utilizes pre-trained off-the-shelf classifiers within the randomized smoothing framework. Rather than directly predicting the label of a noise-augmented image, it first feeds the perturbed image into a \emph{denoiser}, \emph{e.g.}, a diffusion model, and then obtains the predicted label of the denoised image using off-the-shelf pre-trained classifiers that have been trained on clean images. Intriguingly, denoised smoothing with recently developed diffusion models and pre-trained classifiers, \emph{e.g.}, guided diffusion \citep{dhariwal2021diffusion} and BEiT \citep{baobeit}, shows its superior scalability with comparable certified robustness in \(\ell_2\)-adversary to the current state-of-the-art methods \citep{horvath2022robust,jeong2023confidence}.

\begin{figure*}[t]
\centering\small

\includegraphics[width=1.0\textwidth]{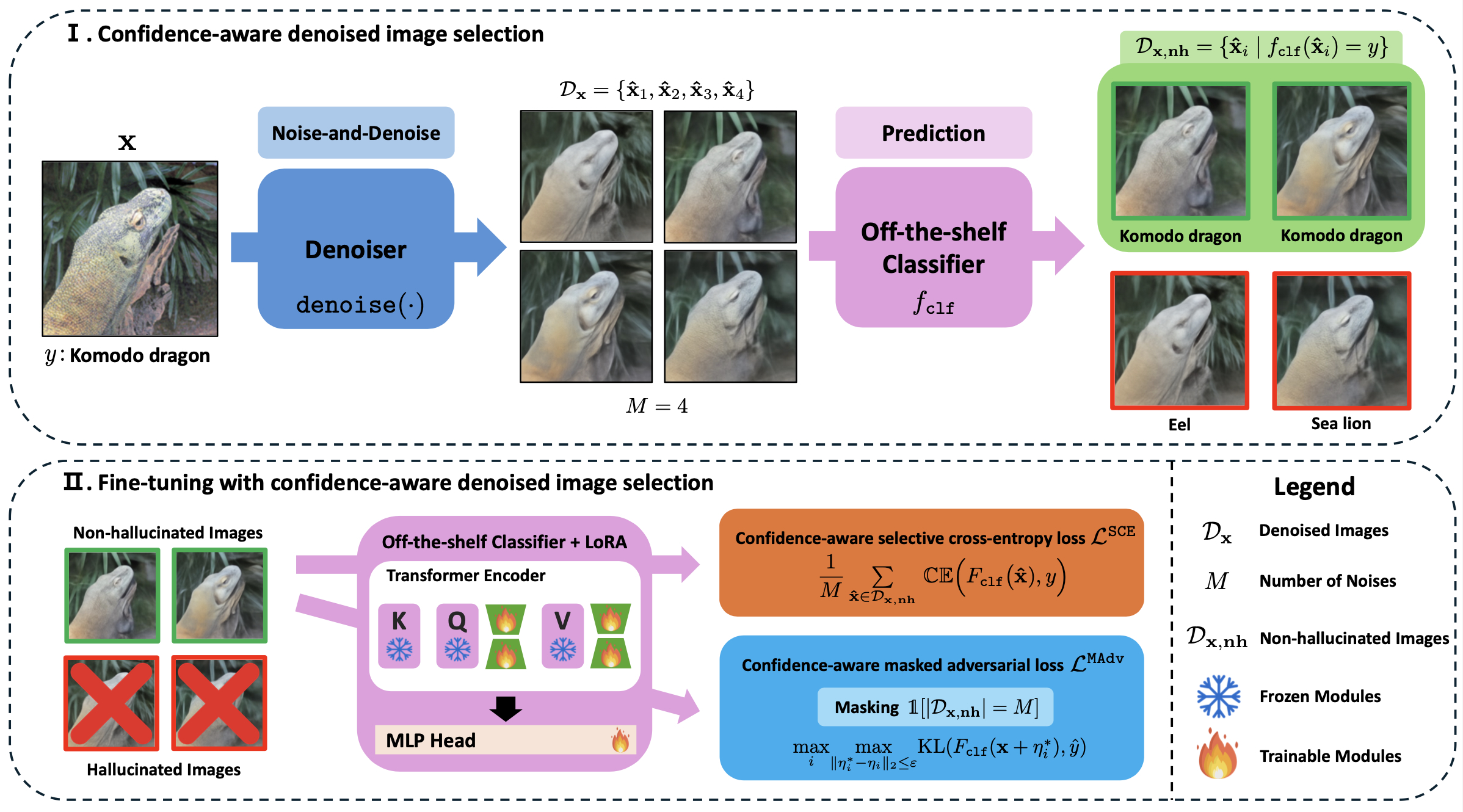}

\caption{Overview of \ALgname framework. (1) Confidence-aware denoised image selection: for a given clean image, we create denoised images and find non-hallucinated images. (2) Fine-tuning with confidence-aware denoised image selection: we propose fine-tuning objectives to improve both generalizability and robustness of the smoothed classifier based on selected non-hallucinated images.}

\label{fig:main_figure}
\end{figure*}

On the other hand, denoised smoothing also exhibits clear limitations. Firstly, denoised images do not follow the standard pre-training distribution of the classifiers, which results in a limited robustness of the denoised smoothing framework. Secondly, fine-tuning the pre-trained classifiers with the denoised images also yields sub-optimal classifiers due to the \emph{hallucinated} images \citep{carlinicertified}, \emph{i.e.}, the diffusion denoiser tends to generate image semantics from an incorrect class rather than the originally assigned class (see Figure~\ref{fig:hallucinated_images}). Consequently, denoised smoothing with such classifiers leads to a drop of the certified accuracy, especially in the large \(\ell_2\)-radius regime, \emph{i.e.}, high Gaussian variance (see Table~\ref{imagenet-table}).

\textbf{Contribution.} In this paper, we aim to address the aforementioned issues of denoised smoothing by designing a fine-tuning objective for off-the-shelf classifiers that distinguishes between \emph{hallucinated} images, \emph{i.e.}, images that have lost the original semantics after denoising, and \emph{non-hallucinated} images, \emph{i.e.}, images that maintain the original semantics after denoising. To this end, we propose to use the ``likelihood of denoised images'', \emph{i.e.}, \emph{confidence}, of the off-the-shelf classifier with respect to the originally assigned class as a proxy for determining whether an image is hallucinated and then fine-tune the classifier with non-hallucinated images only.  
Consequently, we have developed a confidence-aware training objective based on the likelihood of denoised images to effectively discriminate hallucinated images (see Figure \ref{fig:main_figure}).

Specifically, we propose a scalable and practical framework for fine-tuning off-the-shelf classifiers, coined \emph{\Algname} (\ALgname), which improves certified robustness under denoised smoothing. 
In order to achieve this, two new losses are defined: the \emph{\LowLoss} and the \emph{\HighLoss}. Two losses are selectively applied only to non-hallucinated images, thereby ensuring that the overall training process avoids over-optimizing hallucinated samples, \ie samples that are harmful for generalization, while maximizing the robustness of smoothed classifiers.
{Our particular loss design is motivated by} \citet{jeong2023confidence}, {who were the first to investigate training objectives for randomized smoothing depending on sample-wise confidence information. We demonstrate that our novel definition of confidence in randomized smoothing, specifically through the ratio of non-hallucinated images from a denoiser, can dramatically stabilize the confidence-aware training, overcoming its previous limitation of severe accuracy degradation} (\eg see Table~\ref{imagenet-table}). 

In our experiments, we have validated the effectiveness of our proposed method on standard benchmarks for certified \(\ell_2\)-robustness, \emph{i.e.}, CIFAR-10 \citep{alex2009learning} and ImageNet \citep{ILSVRC15}. Our results show that the proposed method significantly outperforms existing state-of-the-art denoised smoothing methods in certified robustness across all \(\ell_2\)-norm setups, while updating only 1\% of the parameters of off-the-shelf classifiers on ImageNet. In particular, \ALgname significantly improves the certified robustness in the high Gaussian variance regime, \emph{i.e.}, high certified radius. For instance, \ALgname outperforms the best performing baseline, \emph{i.e.}, diffusion denoised \citep{carlinicertified}, by 29.5\% $\rightarrow$ 39.4\% at $\varepsilon$ = 2.0 for ImageNet experiments.

\section{Preliminaries}
\label{sec:preliminaries}

\textbf{Adversarial robustness and randomized smoothing.} We assume a labeled dataset $D = \{(\mathbf{x}_i,y_i)\}_{i=1}^n$ sampled from $P$, where $\mathbf{x}_i \in \mathcal{X} \subset \mathbb{R}^d$ and $y_i \in \mathcal{Y} := \{1, ..., K\}$, and aim to develop a classifier $f:\mathcal{X} \rightarrow \mathcal{Y}$ which correctly classifies a given input $\mathbf{x}$ into the corresponding label among $K$ classes, \ie $f(\mathbf{x}_i)=y_i$.

\emph{Adversarial robustness} refers to the \emph{worst-case} behavior of $f$; given a sample $\mathbf{x} \in \mathcal{X}$ and the corresponding label $y \in \mathcal{Y}$, it requires $f$ to produce a consistent output under any perturbation $\delta \in \mathbb{R}^d$ which preserves the original semantic of $\mathbf{x}$. Here, $\delta$ is commonly assumed to be restricted in some \(\ell_2\)-norm in $\mathbb{R}^d$, \emph{i.e.}, $\lVert \delta \rVert_2 \leq \varepsilon$ for some positive $\varepsilon$. For example, \citet{moosavi2016deepfool, carlini2019evaluating} quantify adversarial robustness as \emph{average minimum distance} of the perturbations that cause $f$ to flip the originally assigned label $y$, defined as: 
\begin{align} \label{eqn:adversasrial_robustness}
R(f ; P ) := \mathbb{E}_{\,(\mathbf{x},y) \sim P} \left[ \min_{f(\mathbf{x'}) \neq y} \|\mathbf{x'} - \mathbf{x}\|_2 \right]\,. 
\end{align}

The primary obstacle in achieving adversarial robustness lies in the difficulty of evaluating and optimizing for it, which is typically infeasible because \(f\) is usually modeled by a complex, high-dimensional neural network. \emph{Randomized smoothing} \citep{cohen2019certified, lecuyer2019certified} addresses this challenge by constructing a new robust classifier \(g\) from \(f\), instead of directly modeling robustness with \(f\). In particular, \citet{cohen2019certified} models $g$ by selecting the \emph{{most probable}} output of $f$ under Gaussian noise \(\mathcal{N}(0, \sigma^2 \mI)\), defined as:
\begin{align} \label{eqn:randomized_smoothing_smoothed_classifier}
g(\mathbf{x}) := \underset{c \in \mathcal{Y}}{\arg\max} \,\mathbb{P}_{\delta \sim \mathcal{N}(0, \sigma^2 \mI)} [f(\mathbf{x} + \delta) = c]\,\,.
\end{align}
Intriguingly, $g$ can \emph{guarantee} the adversarial robustness around $(\mathbf{x},y) \sim P$, \emph{i.e.}, $R(g; \mathbf{x}, y)$ can be lower-bounded by the \emph{certified radius} $\underbar{\textit{R}}(g,\mathbf{x}, y)$, where \citet{cohen2019certified} have proven that such a lower-bound of certified radius is tight for \(\ell_2\)-adversary:
\begin{align} \label{eqn:randomized_smoothing_radius}
\textit{R}(g; \mathbf{x}, y) \geq \sigma \cdot \Phi^{-1}(p_{g}(\mathbf{x}, y)) =: \underbar{\textit{R}}(g, \mathbf{x}, y), \quad
\textnormal{where} \quad p_{g}(\mathbf{x}, y) := \mathbb{P}_{\delta}[f(\mathbf{x} + \delta) = y],
\end{align}
provided that \(g(\mathbf{x}) = y\), \emph{i.e.}, $y$ is the {{most probable}} output of $f$ under Gaussian noise. Otherwise, we have \(R(g; \mathbf{x}, y) := 0\). Here, $\Phi$ is the cumulative distribution function of the standard Gaussian distribution. We remark that higher $p_g(\mathbf{x},y)$, \emph{i.e.}, average accuracy of $f(\mathbf{x}+\delta$), results in higher robustness.

\textbf{Denoised smoothing.}
In randomized smoothing, it is crucial that \( f \) consistently classifies perturbed images correctly. \citet{salman2020denoised} have proposed to define \(f\) based on concatenating a Gaussian denoiser, denoted as \texttt{denoise}(\(\cdot\)), with any off-the-shelf classifier \(f_{\texttt{clf}}\), \emph{i.e.}, trained with non-perturbed images, a method referred to as \emph{denoised smoothing}: 
\begin{align} \label{eqn:denoised_smoothing}
f(\mathbf{x} + \delta) := f_{\texttt{clf}}(\texttt{denoise}(\mathbf{x} + \delta))\,\,.
\end{align}

Denoised smoothing provides a more scalable framework for randomized smoothing. First, we only need off-the-shelf pre-trained classifiers (rather than noise-specialized classifiers), which is widely investigated and developed \citep{dosovitskiy2020image, baobeit, radford2021learning}. Second, recent advancements in \emph{diffusion models} \citep{ho2020denoising,nichol2021improved,dhariwal2021diffusion} have produced appropriate denoisers for this approach. Previous efforts \citep{lee2021provable, carlinicertified} have further demonstrated the potential of denoised smoothing in achieving the state-of-the-art certified robustness when combined with recently advanced pre-trained classifiers and diffusion models.

\textbf{Parameter-efficient fine-tuning.}
{LoRA} \citep{hulora} is a widely-used parameter-efficient fine-tuning method that originated from language models. It applies a low-rank constraint to approximate the update matrix at each layer of the Transformer's self-attention layer, significantly reducing the number of trainable parameters for downstream tasks. During fine-tuning, all the parameters of the original model are frozen, and the update of the layer is constrained by representing them with a low-rank decomposition. A forward pass \( h = \mW_0x \) can be modified as follows:
\begin{align} \label{eqn:LoRA}
h = \mW_0x + \Delta \mW x = \mW_0x + \mB\mA x,
\end{align}
where \( x \) and \( h \) denote the input and output features of each layer, \( \mW_0 \in \mathbb{R}^{d \times k} \) represents the original weights of the base model \({f} \), while \( \Delta \mW \) denotes the weight change, composed of the inserted low-rank matrices \( \mB \in \mathbb{R}^{d \times r} \) and \( \mA \in \mathbb{R}^{r \times k} \).

\section{Method}
\label{sec:method}

In Section~\ref{sec:overview}, we present a description of our problem and the main idea. In Section~\ref{sec:method1}, we provide descriptions of our selection strategy for non-hallucinated samples. In Section~\ref{sec:FT-CADIS}, we outline our overall fine-tuning framework.

\subsection{Problem description and Overview}
\label{sec:overview}
\begin{figure}[t]
    \centering
    \begin{minipage}{0.45\textwidth}
        \centering
        \begin{subcaptionblock}{\linewidth}
            \centering
            \includegraphics[width=\linewidth]{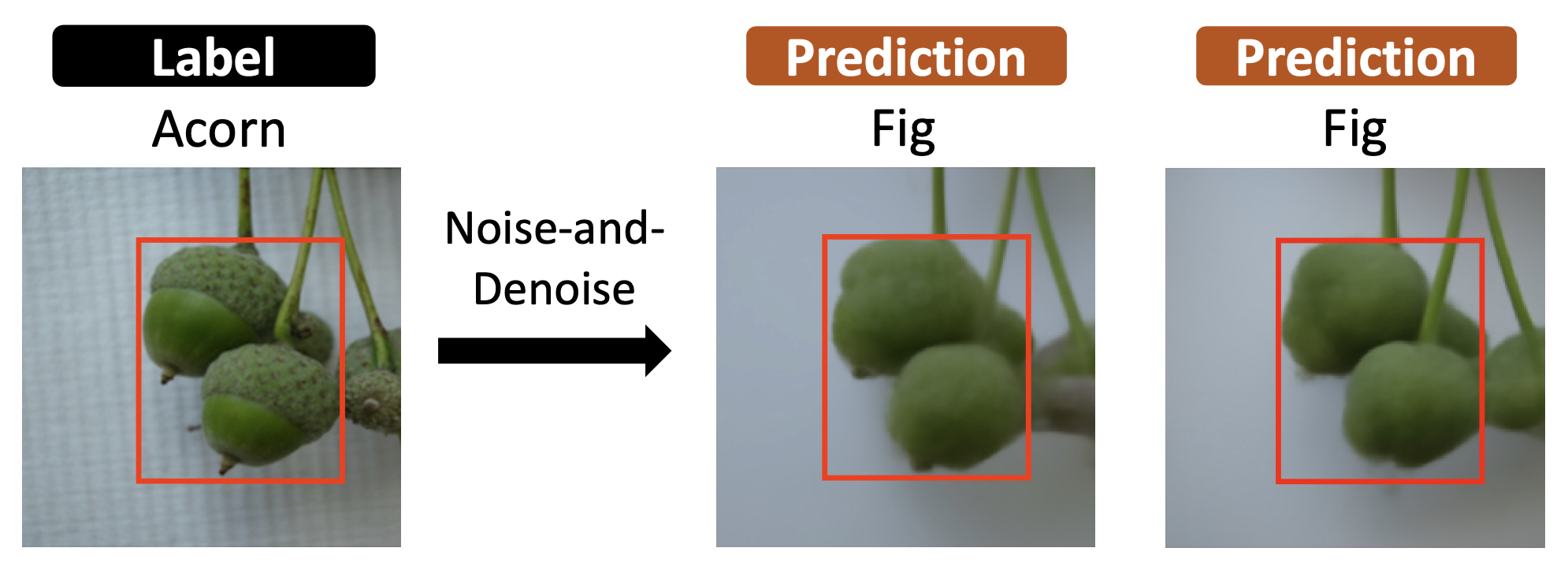}
            \vfill
            \includegraphics[width=\linewidth]{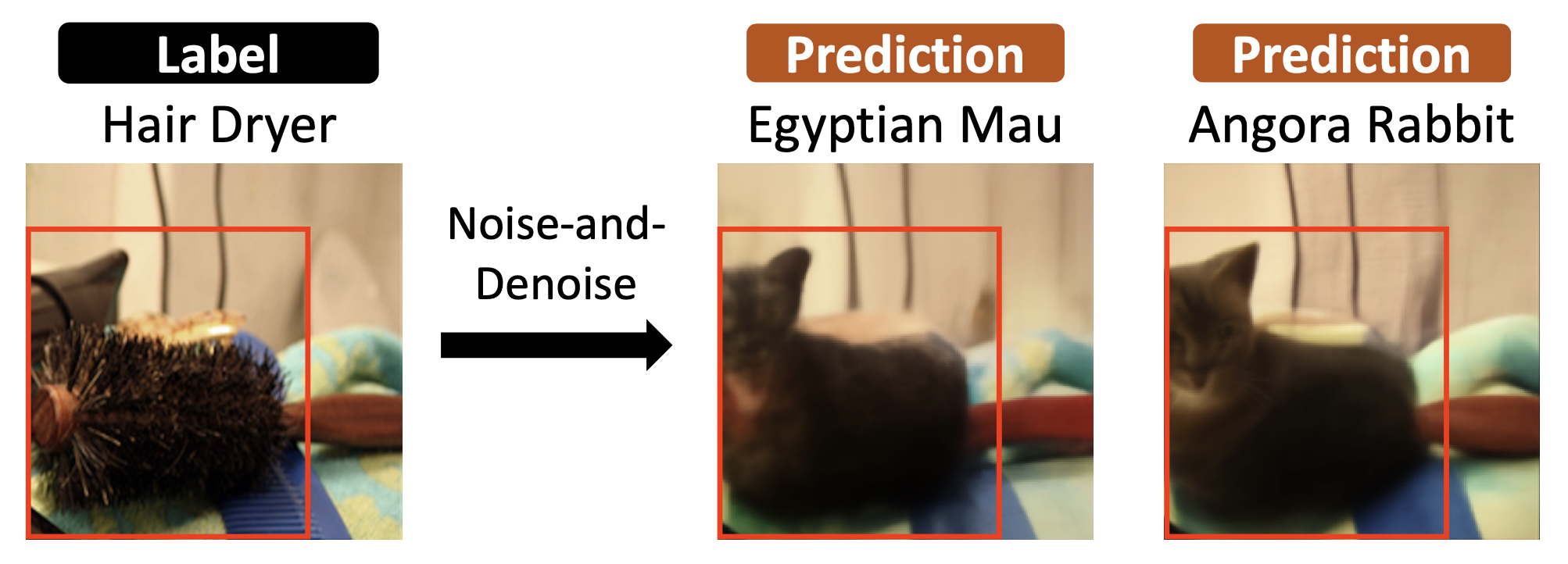}
            \caption{Hallucinated images}
            \label{fig:hallucinated_images}
        \end{subcaptionblock}
    \end{minipage}
    \hfill
    \begin{minipage}{0.45\textwidth}
        \centering
        \begin{subcaptionblock}{\linewidth}
            \centering
            \includegraphics[width=\linewidth]{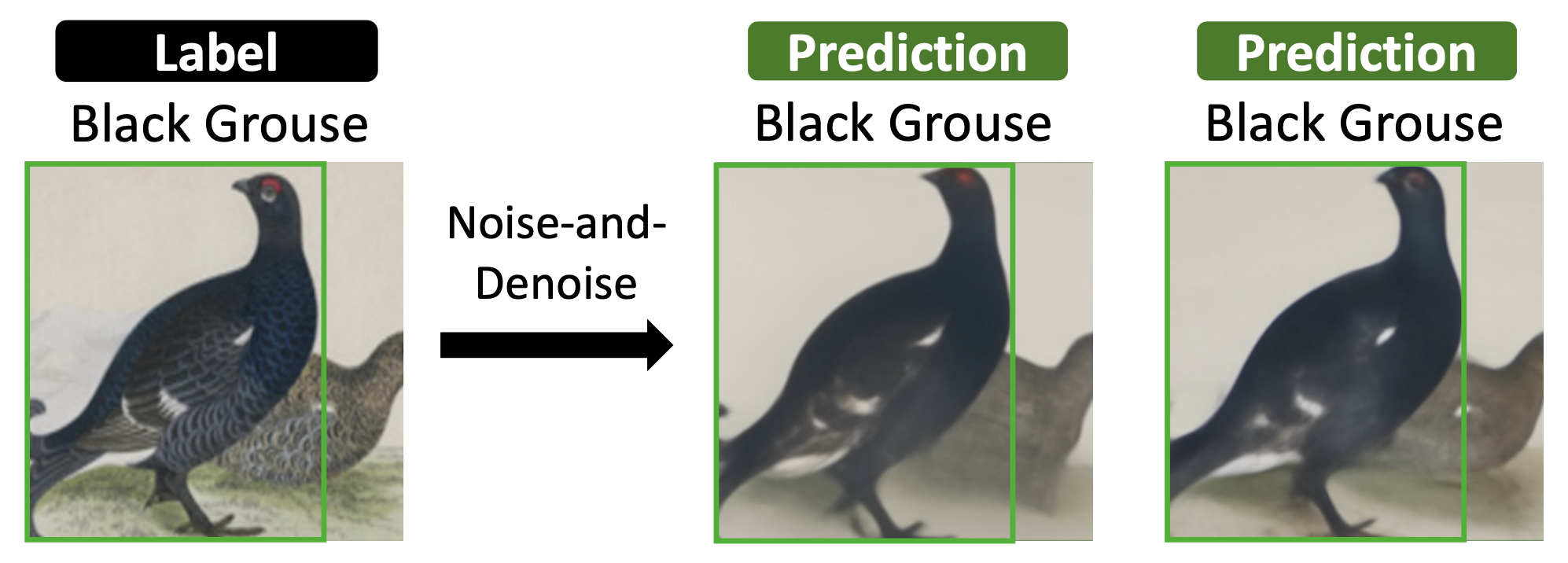}
            \vfill
            \includegraphics[width=\linewidth]{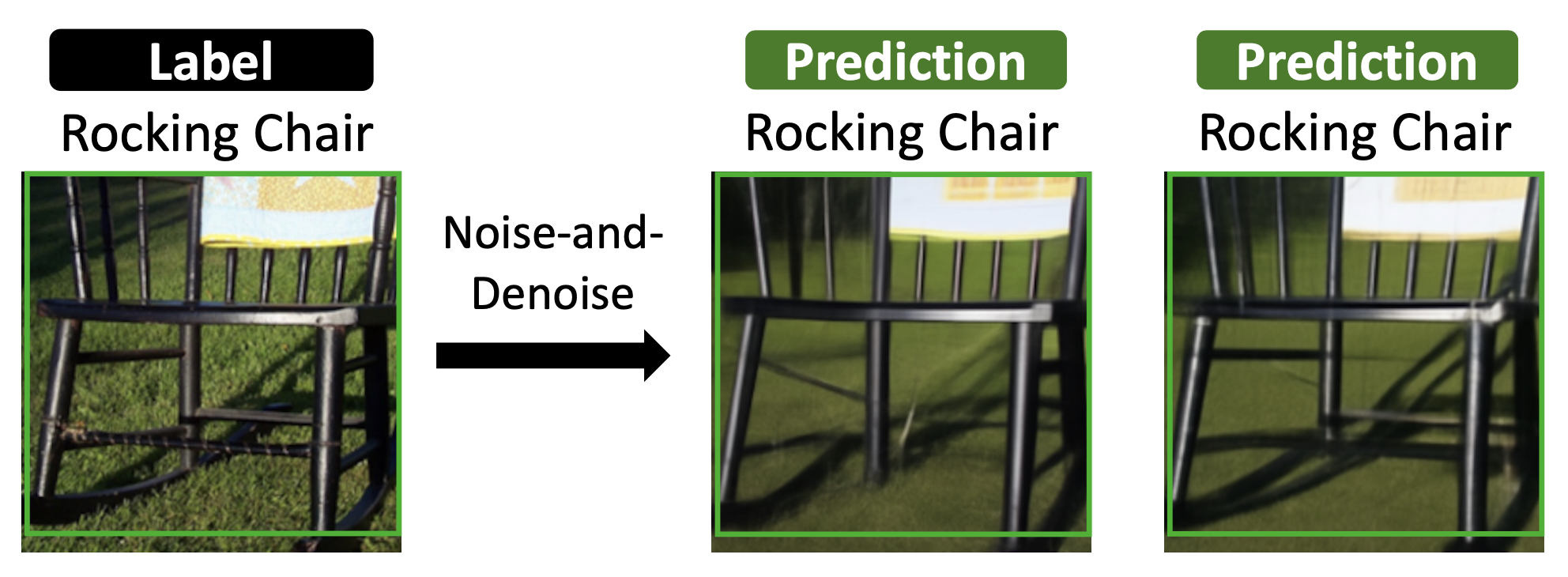}
            \caption{Non-hallucinated images}
            \label{fig:non_hallucinated_images}
        \end{subcaptionblock}
    \end{minipage}
    \caption{Examples of denoised images for \ALgname on ImageNet at $\sigma = 1.00$. We visualize (a) hallucinated images and (b) non-hallucinated images after the noise-and-denoise procedure. The red/green box indicates the areas where the original semantic of the image is corrupted/preserved, respectively.}
    \label{fig:denoised_images}
\end{figure}

In this paper, we investigate \emph{how} to effectively elaborate an off-the-shelf classifier $f_{\texttt{clf}}$ within a denoised smoothing scheme. We remark that the robustness of the smoothed classifier $g$ from denoised smoothing of $f_{\texttt{clf}}$ depends directly on the accuracy of the \emph{denoised} images (see Eq. (\ref{eqn:randomized_smoothing_radius}) and (\ref{eqn:denoised_smoothing})). Therefore, one may expect that improving $f_{\texttt{clf}}$ for clean images is sufficient to improve the generalizability and robustness of $g$ \citep{carlinicertified}, assuming that the denoised images follow the pre-training distribution with clean images \citep{salman2020denoised}, \ie the denoised images preserve the semantics of the original clean images. However, this assumption is not true; the noise-and-denoise procedure of denoised smoothing often suffers from distribution shifts and \emph{hallucination} issues so that the resulting denoised images have completely different semantics from the original labels (see Figure~\ref{fig:hallucinated_images}).

To alleviate these issues, we aim to develop a fine-tuning scheme for $f_{\texttt{clf}}$ to properly handle denoised samples. One straightforward strategy would be to fine-tune $f_{\texttt{clf}}$ by minimizing the cross-entropy loss with \emph{all} denoised images \citep{carlinicertified}:
\begin{align} \label{eqn:gaussian_ce_loss}
\mathcal{L}^{\mathtt{CE}} := \dfrac{1}{M}\displaystyle\sum_{i=1}^M \,\mathbb{CE}\Big({f_{\texttt{clf}}}\big(\texttt{denoise}(\mathbf{x} + \delta_i)\big),\,y\Big),\,\,\delta_i \sim \mathcal{N}(0,\sigma^2\mI),
\end{align}
{{where $\mathbb{CE}$ denotes the cross-entropy loss, and $M$ denotes the number of noises}}. Here, we note that this approach treats both \emph{non-hallucinated} and \emph{hallucinated} samples equally among the denoised samples. However, fine-tuning $f_{\texttt{clf}}$ with hallucinated samples, \ie $\texttt{denoise}(x+\delta_i)$ does not resemble the class $y$, is harmful for the generalizability since Eq. (\ref{eqn:gaussian_ce_loss}) forces the classifier $f_{\texttt{clf}}$ to \emph{remember} non-$y$-like hallucinated images as $y$. Our contribution lies in resolving this issue by introducing (1) a \emph{confidence-aware} selection strategy to distinguish between hallucinated and non-hallucinated images and (2) a fine-tuning strategy that excludes hallucinated samples from the optimization process.

\subsection{Confidence-aware denoised image selection}
\label{sec:method1}

We propose a confidence-aware selection strategy to identify hallucinated images and non-hallucinated images within a set of denoised images. Consider the denoised images $\mathcal{D}_{\mathbf{x}} = \{\texttt{denoise}(\mathbf{x}+\delta_1), ..., \texttt{denoise}(\mathbf{x}+\delta_M)\}$ for a given clean image $\mathbf{x}$ and the number of noises $M$. We aim to find non-hallucinated images within $\mathcal{D}_{\mathbf{x}}$ that an off-the-shelf classifier $f_{\texttt{clf}}$ classifies as the assigned label $y$, \ie ${f_{\texttt{clf}}}$ shows the highest confidence for $y$ among all possible classes. Conversely, if $f_{\texttt{clf}}$ classifies denoised images as a label other than $y$, we define such denoised images as hallucinated images, \ie samples that no longer preserve the core semantic of $y$. Accordingly, the set of non-hallucinated images $\mathcal{D}_{\mathbf{x, nh}} \in \mathcal{D}_{\mathbf{x}}$ is defined as follows:
\begin{align} \label{eqn:non_hallucinated_set}
\mathcal{D}_{\mathbf{x, nh}} &= \{\mathbf{\hat{x}} \vert f_{\texttt{clf}}(\texttt{denoise}(\mathbf{x}+\delta_i)) = y, \,i \in [1,...,M] \}\,\,.
\end{align}

We remark that the off-the-shelf classifier $f_{\texttt{clf}}$ is pre-trained with clean images, rather than denoised images. Thus, at the beginning of the fine-tuning, $f_{\texttt{clf}}$ often fails to correctly assign $\mathcal{D}_{\mathbf{x, nh}}$ due to the distribution shift from clean images to denoised images. Thus, we update $\mathcal{D}_{\mathbf{x, nh}}$ at each training iteration using Eq. (\ref{eqn:non_hallucinated_set}) for a more accurate assignment of non-hallucinated images.

\subsection{Fine-tuning with confidence-aware denoised image selection}
\label{sec:FT-CADIS}

Our main goal is to improve both the generalizability and the robustness of the smoothed classifier $g$, through the fine-tuning of the off-the-shelf classifier $f_{\texttt{clf}}$ based on our confidence-aware denoised image selection in Section~\ref{sec:method1}. To this end, we propose two fine-tuning objectives for an off-the-shelf classifier $f_{\texttt{clf}}$, \viz \LowLoss and \HighLoss, to maximize the generalizability and robustness of the corresponding smoothed classifier $g$, respectively.

\textbf{\LowLoss.} We first aim to improve the generalizability of the smoothed classifier $g$, \ie the average certified accuracy of $g$. Specifically, we propose to optimize $f_{\texttt{clf}}$ with non-hallucinated images $\mathcal{D}_{\mathbf{x, nh}}$:

\begin{align} \label{eqn:ce_nh_loss}
\mathcal{L}^{\mathtt{SCE}} := \dfrac{1}{M} \sum\limits_{\smash{\mathbf{\hat{x}} \in \mathcal{D}_{\mathbf{x, nh}}}} \,\mathbb{CE}\Big({f_{\texttt{clf}}}\big(\mathbf{\hat{x}}), \,y\Big)\,\,.
\end{align}

In other words, we optimize our classifier with the non-hallucinated images, while the hallucinated images are excluded from our training objective. This prevents the drop in accuracy of $f_{\texttt{clf}}$ caused by being forced to remember wrong semantics not relevant to the assigned class $y$. It also allows for $f_{\texttt{clf}}$ to properly learn the distribution of \emph{denoised} images, which is largely different from its pre-training distribution with clean images.

Here, we find that training with the objective in Eq. (\ref{eqn:ce_nh_loss}) slows down the overall training procedure since $\mathcal{D}_{\mathbf{x, nh}} = \emptyset$ sometimes occurs at the start of training. This is mainly due to the distribution shift from the pre-training clean image distribution to the denoised images, \ie $f_{\texttt{clf}}$ fails to classify denoised images due to insufficient exposure to denoised images. To resolve this \emph{cold-start} problem, we add the most-$y$-like denoised image, \ie a denoised image with the largest logit for $y$, to $\mathcal{D}_{\mathbf{x, nh}}$ when it is empty.

\textbf{\HighLoss.} We also propose a simple strategy to further improve the robustness of the smoothed classifier $g$, \ie the certified accuracy of $g$ at large \(\ell_2\)-norm radius. Specifically, we apply the concept of \emph{adversarial training} \citep{madry2018towards, zhang2019theoretically, wang2019improving, salman2019provably, jeong2023confidence} to our denoised smoothing setup; we carefully create {more challenging} images, and then additionally learn these images during fine-tuning. Here, the main challenge is to ensure that the adversarial images \emph{preserve} the core semantic of the original image, thereby maintaining generalizability while improving robustness. However, as illustrated in Figure~\ref{fig:denoised_images}, some clean images are prone to be hallucinated after the noise-and-denoise procedure. Therefore, adversarial training in denoised smoothing should be carefully designed to avoid incorporating hallucinated images.

To this end, we propose to create adversarial examples based only on images that are unlikely to be hallucinated, \ie clean images $\mathbf{x}$  with $\mathcal{D}_{\mathbf{x, nh}} = \mathcal{D}_{\mathbf{x}}$. Specifically, we apply our adversarial loss based on a simple condition of ``$\mathcal{D}_{\mathbf{x, nh}} = M$'':
\begin{align} \label{eqn:adv_nh_loss}
\mathcal{L}^{\mathtt{MAdv}} := \mathds{1} {[{|\mathcal{D}_{\mathbf{x, nh}}|=M}]} \cdot \underset{i}{\mathrm{max}}\underset{\lVert \eta_i^{*} - \eta_{i} \rVert_2 \leq \varepsilon}{\mathrm{max}} \textnormal{KL}({f_{\texttt{clf}}}(\mathbf{x}+\eta_i^{*}), \,\hat{y}),
\end{align}

where \textnormal{KL($\cdot,\cdot$)} indicates the Kullback-Libler divergence and $\eta_i := \texttt{denoise}(\mathbf{x} + \delta_i) - \mathbf{x}$ is the difference between each denoised image and the original clean image. To find the adversarial perturbation $\eta_i^*$, we perform a $T$-step gradient ascent from each $\eta_i$ with a step size of $2 \cdot \varepsilon / T$, while projecting $\eta_i^*$ to remain within an $\ell_2$-ball of radius $\varepsilon$: \viz the \emph{projected gradient descent} (PGD) \citep{madry2018towards}. For the adversarial target $\hat{y}$, we adapt the \emph{consistency} target from the previous robust training method \citep{jeong2023confidence} to our denoised smoothing setup by letting the target be the average likelihood of the denoised images, \ie $\hat{y}:=\frac{1}{M}\sum^{M}_{i=1} \,\texttt{Softmax}\big({f_{\texttt{clf}}}\big(\texttt{denoise}(\mathbf{x} + \delta_i)\big)\big)$.

\textbf{Overall training objective.} Building on our proposed training objectives $\mathcal{L}^{\mathtt{SCE}}$ and $\mathcal{L}^{\mathtt{MAdv}}$, we now present the complete objective for our \emph{\Algname} (\ALgname). Based on our confidence-aware denoised image selection scheme, \LowLoss and \HighLoss are applied only to non-hallucinated images $\mathcal{D}_{\mathbf{x, nh}}$ to improve both generalizability and robustness of the smoothed classifier. 
The overall loss function is as follows:
\begin{align} \label{eqn:overall_objective}
\mathcal{L}^{\mathtt{\ALgname}} := \mathcal{L}^{\mathtt{SCE}} + \lambda \cdot \mathcal{L}^{\mathtt{MAdv}}, 
\end{align}
where \(\lambda\) > 0 is a hyperparameter, which controls the relative trade-off between the generalizability and the robustness (see Section~\ref{subsec:ablation}). 
The detailed algorithm for computing our $\mathcal{L}^{\mathtt{\ALgname}}$ is outlined in Algorithm \ref{appendix:alg:FT-CADIS}.

\textbf{Comparision with CAT-RS.} Our \ALgname has drawn motivation from previous confidence-aware training strategies, \eg CAT-RS \citep{jeong2023confidence}. The key difference is that \ALgname uses the confidence of denoised images based on the pre-trained off-the-shelf classifier while CAT-RS learns their confidence of Gaussian-augmented images during the training of the classifier from scratch. In particular, our method takes advantage of off-the-shelf classifiers which are already capable of providing reasonable confidence for identifying non-hallucinated images. Therefore, we can simply use the non-hallucinated images identified by the off-the-shelf classifiers in our optimization objective. On the other hand, CAT-RS additionally assumes a distribution of semantic-preserving noised sample counts based on the confidence, $\ie$ average accuracy, of the models currently being trained from scratch. Therefore, the overall confidence remains low especially for complex datasets, resulting in a sub-optimal accuracy of the smoothed classifier (see Table~\ref{imagenet-table}). Our \ALgname successfully mitigates this issue based on our carefully designed confidence-based approach utilizing off-the-shelf classifiers, achieving the state-of-the-art robustness even in complex datasets such as ImageNet.

\section{Experiments}
\label{sec:experiments}

We verify the effectiveness of our proposed training scheme for off-the-shelf classifiers by conducting comprehensive experiments. In Section~\ref{subsec:experimental_setup}, we explain our experimental setups, such as training configurations and evaluation metrics. In Section~\ref{subsec:main_experiments}, we present the main results on CIFAR-10 and ImageNet. In Section~\ref{subsec:ablation}, we conduct an ablation study to analysis the component-wise effect of our training objective.

\subsection{Experimental setup}
\label{subsec:experimental_setup}

\textbf{Baselines.} We mainly consider the following recently proposed methods based on \emph{denoised smoothing} \citep{salman2020denoised,lee2021provable,carlinicertified,jeong2024multi} framework. We additionally compare with other robust training methods for certified robustness based on randomized smoothing \citep{lecuyer2019certified,cohen2019certified,salman2019provably,jeong2020consistency,zhaimacer,horvathboosting,yang2022certified,jeong2021smoothmix,horvath2022robust,jeong2023confidence}. Following the previous works, we consider three different noise levels, \(\sigma \in \{0.25, 0.50, 1.00\}\), to obtain smoothed classifiers.

{\bf CIFAR-10 configuration.} We follow the same classifier and the same denoiser employed by \citet{carlinicertified}. Specifically, we use the 86M-parameter ViT-B/16 classifier \citep{dosovitskiy2020image} which is pre-trained and fine-tuned on ImageNet-21K \citep{deng2009imagenet} and CIFAR-10 \citep{alex2009learning}, respectively. We use the 50M-parameter 32$\times$32 diffusion model from \citet{nichol2021improved} as the denoiser. We provide more detailed setups in Appendix \ref{appendix:Training}.

\textbf{ImageNet configuration.} We use the 87M-parameter ViT-B/16 classifier which is pre-trained on LAION-2B image-text pairs \citep{schuhmann2022laion} using OpenCLIP \citep{cherti2023reproducible} and fine-tuned on ImageNet-12K and then ImageNet-1K. Compared to the previous state-of-the-art method, diffusion denoised \citep{carlinicertified} based on BEiT-large model \citep{baobeit} with 305M parameters, we use a much smaller off-the-shelf classifier (30\% parameters). We also adopt parameter-efficient fine-tuning with LoRA \citep{hulora}, \ie the number of parameters updated through fine-tuning is only 1\% of the total parameters. We use the same denoiser employed by \citet{carlinicertified}, \ie 552M-parameter 256$
\times$256 unconditional model from \citet{dhariwal2021diffusion}. We provide more detailed setups in Appendix \ref{appendix:Training}.

{\bf Evaluation metrics.} We follow the standard metric in the literature for assessing the certified robustness of smoothed classifiers : the \emph{approximate certified test accuracy} at \emph{r}, which is the fraction of the test set that \textsc{Certify} \citep{cohen2019certified}, a practical Monte-Carlo-based certification procedure, classifies correctly with a radius larger than \emph{r} without abstaining. Throughout our experiments, following 
\citet{carlinicertified}, we use  $N = 100,000$  noise samples to certify robustness for entire CIFAR-10 test set and  $N = 10,000$ samples for 1,000 randomly selected images from the ImageNet validation set (note that \emph{RS} methods in Table \ref{imagenet-table} use $N = 100,000$). We use the hyperparameters from \citet{cohen2019certified}, specifically $n_0 = 100$  and  $\alpha$ = 0.001. In ablation study, we additionally consider another standard metric, the \emph{average cerified radius} (ACR) \citep{zhaimacer}: the average of cerified radii on the test set $D_{test}$ while assigning incorrect samples as 0: \viz $\mathrm{ACR} := \frac{1}{|D_{test}|}\sum_{(\mathbf{x},y) \in D_{test}}[\mathrm{CR}(f,\sigma,\mathbf{x}) \cdot \mathds{1}_{g(\mathbf{x})=y}],$ where $\mathrm{CR}(\cdot)$ denotes the lower bound of certified radius \textsc{Certify} returns.

\begin{table}[t]
\centering
\caption{CIFAR-10 and ImageNet certified top-1 accuracy. We report the best certified accuracy among the models trained with $\sigma \in \{0.25, 0.50, 1.00\}$, followed by the clean accuracy of the corresponding model in parentheses. RS denotes methods based on randomized smoothing without a denoising procedure, and DS denotes methods based on denoised smoothing. \emptycirc[1ex] indicates training the classifier with Gaussian-augmented images, \fullcirc[1ex] indicates direct use of the off-the-shelf classifier without fine-tuning, \halfcirc[1ex] indicates fine-tuning of the denoiser, \halfcircc[1ex] indicates fine-tuning the off-the-shelf classifier, and \HalfHatchedCircle[1ex] indicates parameter-efficient fine-tuning of the off-the-shelf classifier \citep{hulora}. The highest certified accuracy in each column is bold-faced. $\dagger$ indicates that extra data is used in the pre-training.}
\label{main-experiments-table} 
\begin{tabular}{c}
    \begin{subcaptionblock}{0.95\textwidth}
    \caption{CIFAR-10}
    \label{cifar-10-table}
    \centering
    \resizebox{\textwidth}{!}{%
    \begin{tabular}{clcccccccc}
    \toprule
        \multirow{2.5}{*}{{Category}} & 
        \multirow{2.5}{*}{{Method}} & 
        \multirow{2.5}{*}{\shortstack{Off-the-shelf}} &
        \multicolumn{6}{c}{Certified Accuracy at $\varepsilon$ (\%)} \\ \cmidrule{4-9}
        &&& {0.25} & {0.50} & {0.75} & {1.00} & {1.25} & {1.50} 
        \\
        \toprule
        \multirow{10}{*}{\emph{{RS}}} & PixelDP \citep{lecuyer2019certified} & \emptycirc \emptycirc[0ex] & \( {\text{\small (71.0)}}_{\text{\large 22.0}} \) & \( {\text{\small (44.0)}}_{\text{\large 2.0}} \) & - & - & - & - \\
        & Gaussian \citep{cohen2019certified} & \emptycirc \emptycirc[0ex] & \( {\text{\small (77.0)}}_{\text{\large 61.0}} \) & \( {\text{\small (66.0)}}_{\text{\large 43.0}} \) & \( {\text{\small (66.0)}}_{\text{\large 32.0}} \) & \( {\text{\small (66.0)}}_{\text{\large 22.0}} \) & \( {\text{\small (47.0)}}_{\text{\large 17.0}} \) & \( {\text{\small (47.0)}}_{\text{\large 14.0}} \)\\ 
        & SmoothAdv \citep{salman2019provably} & \emptycirc \emptycirc[0ex] & \( {\text{\small (85.0)}}_{\text{\large 73.0}} \) & \( {\text{\small (76.0)}}_{\text{\large 58.0}} \) & \( {\text{\small (75.0)}}_{\text{\large 48.0}} \) & \( {\text{\small (57.0)}}_{\text{\large 38.0}} \) & \( {\text{\small (53.0)}}_{\text{\large 33.0}} \) & \( {\text{\small (53.0)}}_{\text{\large 29.0}} \)\\ 
        & Consistency \citep{jeong2020consistency} & \emptycirc \emptycirc[0ex] & \( {\text{\small (77.8)}}_{\text{\large 68.8}} \) & \( {\text{\small (75.8)}}_{\text{\large 58.1}} \) & \( {\text{\small (72.9)}}_{\text{\large 48.5}} \) & \( {\text{\small (52.3)}}_{\text{\large 37.8}} \) & \( {\text{\small (52.3)}}_{\text{\large 33.9}} \) & \( {\text{\small (52.3)}}_{\text{\large 29.9}} \)\\ 
        & MACER \citep{zhaimacer} & \emptycirc \emptycirc[0ex] & \( {\text{\small (81.0)}}_{\text{\large 71.0}} \) & \( {\text{\small (81.0)}}_{\text{\large 59.0}} \) & \( {\text{\small (66.0)}}_{\text{\large 46.0}} \) & \( {\text{\small (66.0)}}_{\text{\large 38.0}} \) & \( {\text{\small (66.0)}}_{\text{\large 29.0}} \) & \( {\text{\small (45.0)}}_{\text{\large 25.0}} \)\\ 
        & Boosting \citep{horvathboosting} & \emptycirc \emptycirc[0ex] & \( {\text{\small (83.4)}}_{\text{\large 70.6}} \) & \( {\text{\small (76.8)}}_{\text{\large 60.4}} \) & \( {\text{\small (71.6)}}_{\text{\large 52.4}} \) & \( {\text{\small (52.4)}}_{\text{\large 38.8}} \) & \( {\text{\small (52.4)}}_{\text{\large 34.4}} \)& \( {\text{\small (52.4)}}_{\text{\large \textbf{30.4}}} \)\\ 
        & DRT \citep{yang2022certified} & \emptycirc \emptycirc[0ex] & \( {\text{\small (81.5)}}_{\text{\large 70.4}} \) & \( {\text{\small (72.6)}}_{\text{\large 60.2}} \) & \( {\text{\small (71.9)}}_{\text{\large 50.5}} \) & \( {\text{\small (56.1)}}_{\text{\large 39.8}} \) & \( {\text{\small (56.4)}}_{\text{\large \textbf{36.0}}} \) & \( {\text{\small (56.4)}}_{\text{\large \textbf{30.4}}} \) \\ 
        & SmoothMix \citep{jeong2021smoothmix} & \emptycirc \emptycirc[0ex] & \( {\text{\small (77.1)}}_{\text{\large 67.9}} \) & \( {\text{\small (77.1)}}_{\text{\large 57.9}} \) & \( {\text{\small (74.2)}}_{\text{\large 47.7}} \) & \( {\text{\small (61.8)}}_{\text{\large 37.2}} \) & \( {\text{\small (61.8)}}_{\text{\large 31.7}} \) & \( {\text{\small (61.8)}}_{\text{\large 25.7}} \)\\ 
        & ACES \citep{horvath2022robust} & \emptycirc \emptycirc[0ex] & \( {\text{\small (77.6)}}_{\text{\large 69.0}} \) & \( {\text{\small (73.4)}}_{\text{\large 57.2}} \) & \( {\text{\small (73.4)}}_{\text{\large 47.0}} \) & \( {\text{\small (57.0)}}_{\text{\large 37.8}} \) & 
        \( {\text{\small (57.0)}}_{\text{\large 32.2}} \) & 
        \( {\text{\small (57.0)}}_{\text{\large 27.8}} \)\\ 
        & CAT-RS \citep{jeong2023confidence} & \emptycirc \emptycirc[0ex] & 
        \( {\text{\small (76.3)}}_{\text{\large 68.1}} \) & 
        \( {\text{\small (76.3)}}_{\text{\large 58.8}} \) & 
        \( {\text{\small (76.3)}}_{\text{\large 48.2}} \) & 
        \( {\text{\small (62.3)}}_{\text{\large 38.5}} \) & 
        \( {\text{\small (62.3)}}_{\text{\large 32.7}} \) & 
        \( {\text{\small (62.3)}}_{\text{\large 27.1}} \)\\ 
        \midrule
        \multirow{6.5}{*}{\emph{{DS}}}
        & Denoised
        \citep{salman2020denoised} & \halfcirc \halfcirc[0ex] & 
        \( {\text{\small (72.0)}}_{\text{\large 56.0}} \) & 
        \( {\text{\small (62.0)}}_{\text{\large 41.0}} \) & 
        \( {\text{\small (62.0)}}_{\text{\large 28.0}} \) & 
        \( {\text{\small (44.0)}}_{\text{\large 19.0}} \) & 
        \( {\text{\small (42.0)}}_{\text{\large 16.0}} \) & 
        \( {\text{\small (44.0)}}_{\text{\large 13.0}} \)\\ 
        & Score-based Denoised \citep{lee2021provable} & \fullcirc \fullcirc[0ex] & 
        \({\text{\small\phantom{(00.0)}}}_{\text{\large 60.0}} \) & 
        \( {\text{\small\phantom{(00.0)}}}_{\text{\large 42.0}} \) & 
        \( {\text{\small\phantom{(00.0)}}}_{\text{\large 28.0}} \) & 
        \( {\text{\small\phantom{(00.0)}}}_{\text{\large 19.0}} \) & 
        \( {\text{\small\phantom{(00.0)}}}_{\text{\large 11.0}} \) & 
        \( {\text{\small\phantom{(00.0)}}}_{\text{\large 6.0}} \) \\ 
        & Diffusion Denoised{$^\dagger$} \citep{carlinicertified} & \fullcirc \fullcirc[0ex] & 
        \( {\text{\small (88.1)}}_{\text{\large 76.7}} \) & 
        \( {\text{\small (88.1)}}_{\text{\large 63.0}} \) & 
        \( {\text{\small (88.1)}}_{\text{\large 45.3}} \) & 
        \( {\text{\small (77.0)}}_{\text{\large 32.1}} \) 
        & - & - \\
        & Diffusion Denoised{$^\dagger$}\tablefootnote{Further fine-tune the classifier on denoised images from CIFAR-10.} \citep{carlinicertified} & 
        \halfcircc \halfcircc[0ex] & 
        \( {\text{\small (91.2)}}_{\text{\large 79.3}} \) & 
        \( {\text{\small (91.2)}}_{\text{\large 65.5}} \) & 
        \( {\text{\small (91.2)}}_{\text{\large 48.7}} \) & 
        \( {\text{\small (81.5)}}_{\text{\large 35.5}} \) & - & - \\
        & Multi-scale Denoised{$^\dagger$} \citep{jeong2024multi} 
        & \halfcirc \halfcirc[0ex] 
        & - & 
        \( {\text{\small (90.3)}}_{\text{\large 61.9}} \) 
        & - & 
        \( {\text{\small (85.1)}}_{\text{\large 32.9}} \) 
        & - &
        \( {\text{\small (79.6)}}_{\text{\large 16.2}} \) \\
        \cmidrule{2-9}
        & \textbf{\ALgname (Ours)}{$^\dagger$} 
        & \halfcircc \halfcircc[0ex]
        & \( {\text{\small (88.7)}}_{\text{\large \textbf{80.3}}} \) 
        & \( {\text{\small (88.7)}}_{\text{\large \textbf{68.4}}} \) 
        & \( {\text{\small (88.7)}}_{\text{\large \textbf{54.5}}} \) 
        & \( {\text{\small (74.9)}}_{\text{\large \textbf{39.9}}} \) 
        & \( {\text{\small (74.9)}}_{\text{\large 31.6}} \) 
        & \( {\text{\small (74.9)}}_{\text{\large 23.5}} \) \\
        \bottomrule
        \end{tabular}
        }
        \end{subcaptionblock}

        \\
        
        \begin{subcaptionblock}{0.95\textwidth}
        \caption{ImageNet}
        \label{imagenet-table}
        \centering
        \resizebox{\textwidth}{!}{%
        \begin{tabular}{clcccccc}
        \toprule
        \multirow{2.5}{*}{{Category}} & \multirow{2.5}{*}{{Method}} & \multirow{2.5}{*}{{Off-the-shelf}} & \multicolumn{5}{c}{Certified Accuracy at $\varepsilon$ (\%)} \\ \cmidrule{4-8}
         &&& {0.50} & {1.00} & {1.50} & {2.00} & 
        {2.50} 
        \\
        \toprule
        \multirow{10}{*}{\emph{{RS}}} & PixelDP \citep{lecuyer2019certified} & \emptycirc \emptycirc[0ex] &  \( {\text{\small (33.0)}}_{\text{\large 16.0}} \) & - & - & - & - 
        \\
        & Gaussian \citep{cohen2019certified} & \emptycirc \emptycirc[0ex] & \( {\text{\small (67.0)}}_{\text{\large 49.0}} \) & \( {\text{\small (57.0)}}_{\text{\large 37.0}} \) & \( {\text{\small (57.0)}}_{\text{\large 29.0}} \) & \( {\text{\small (44.0)}}_{\text{\large 19.0}} \) & \( {\text{\small (44.0)}}_{\text{\large 15.0}} \) 
        \\ 
        & SmoothAdv \citep{salman2019provably} & \emptycirc \emptycirc[0ex] & \( {\text{\small (65.0)}}_{\text{\large 56.0}} \) & \( {\text{\small (55.0)}}_{\text{\large 45.0}} \) & \( {\text{\small (55.0)}}_{\text{\large 38.0}} \) & \( {\text{\small (42.0)}}_{\text{\large 28.0}} \) & \( {\text{\small (42.0)}}_{\text{\large 26.0}} \) 
        \\ 
        & Consistency \citep{jeong2020consistency} & \emptycirc \emptycirc[0ex] & \( {\text{\small (55.0)}}_{\text{\large 50.0}} \) & \( {\text{\small (55.0)}}_{\text{\large 44.0}} \) & \( {\text{\small (55.0)}}_{\text{\large 34.0}} \) & \( {\text{\small (41.0)}}_{\text{\large 24.0}} \) & \( {\text{\small (41.0)}}_{\text{\large 21.0}} \) 
        \\ 
        & MACER \citep{zhaimacer} & \emptycirc \emptycirc[0ex] &  \( {\text{\small (68.0)}}_{\text{\large 57.0}} \) & \( {\text{\small (64.0)}}_{\text{\large 43.0}} \) & \( {\text{\small (64.0)}}_{\text{\large 31.0}} \) & \( {\text{\small (48.0)}}_{\text{\large 25.0}} \) & \( {\text{\small (48.0)}}_{\text{\large 18.0}} \) 
        \\ 
        & Boosting \citep{horvathboosting} & \emptycirc \emptycirc[0ex] & \( {\text{\small (68.0)}}_{\text{\large 57.0}} \) & \( {\text{\small (57.0)}}_{\text{\large 44.6}} \) & \( {\text{\small (57.0)}}_{\text{\large 38.4}} \) & \( {\text{\small (44.6)}}_{\text{\large 28.6}} \) & \( {\text{\small (38.6)}}_{\text{\large 24.6}} \)
        \\ 
        & DRT \citep{yang2022certified} & \emptycirc \emptycirc[0ex] & \( {\text{\small (52.2)}}_{\text{\large 46.8}} \) & \( {\text{\small (49.8)}}_{\text{\large 44.4}} \) & \( {\text{\small (49.8)}}_{\text{\large 39.8}} \) & \( {\text{\small (49.8)}}_{\text{\large 30.4}} \) & \( {\text{\small (49.8)}}_{\text{\large 29.0}} \) 
        \\ 
        & SmoothMix \citep{jeong2021smoothmix} & \emptycirc \emptycirc[0ex] & \( {\text{\small (55.0)}}_{\text{\large 50.0}} \) & \( {\text{\small (55.0)}}_{\text{\large 43.0}} \) & \( {\text{\small (55.0)}}_{\text{\large 38.0}} \) & \( {\text{\small (40.0)}}_{\text{\large 26.0}} \) & \( {\text{\small (40.0)}}_{\text{\large 24.0}} \) 
        \\ 
        & ACES \citep{horvath2022robust} & \emptycirc \emptycirc[0ex] & \( {\text{\small (63.2)}}_{\text{\large 54.0}} \) & \( {\text{\small (55.4)}}_{\text{\large 42.2}} \) & \( {\text{\small (55.0)}}_{\text{\large 35.6}} \) & \( {\text{\small (39.2)}}_{\text{\large 25.6}} \) & \( {\text{\small (50.6)}}_{\text{\large 22.0}} \) 
        \\ 
        & CAT-RS \citep{jeong2023confidence} & \emptycirc \emptycirc[0ex] & \( {\text{\small (44.0)}}_{\text{\large 38.0}} \) & \( {\text{\small (44.0)}}_{\text{\large 35.0}} \) & \( {\text{\small (44.0)}}_{\text{\large 31.0}} \) & \( {\text{\small (44.0)}}_{\text{\large 27.0}} \) & \( {\text{\small (44.0)}}_{\text{\large 24.0}} \) 
        \\ 
        \midrule
        \multirow{5.5}{*}{\emph{{DS}}}
        & Denoised \citep{salman2020denoised} & \halfcirc \halfcirc[0ex] & \( {\text{\small (60.0)}}_{\text{\large 33.0}} \) & \( {\text{\small (38.0)}}_{\text{\large 14.0}} \) & \( {\text{\small (38.0)}}_{\text{\large 6.0}} \) & - & -  
        \\ 
        & Score-based Denoised \citep{lee2021provable} & \fullcirc \fullcirc[0ex] & \({\text{\small\phantom{(00.0)}}}_{\text{\large 41.0}} \) & \( {\text{\small\phantom{(00.0)}}}_{\text{\large 24.0}} \) & \( {\text{\small\phantom{(00.0)}}}_{\text{\large 11.0}} \) & - & -  
        \\
        & Diffusion Denoised{$^\dagger$}\citep{carlinicertified} & \fullcirc \fullcirc[0ex] & \( {\text{\small (82.8)}}_{\text{\large 71.1}} \) & \( {\text{\small (77.1)}}_{\text{\large 54.3}} \) & \( {\text{\small (77.1)}}_{\text{\large 38.1}} \) & \( {\text{\small (60.0)}}_{\text{\large 29.5}} \) & - 
        \\
        & Multi-scale Denoised{$^\dagger$} \citep{jeong2024multi} & \halfcirc \halfcirc[0ex] & \( {\text{\small (76.6)}}_{\text{\large 54.6}} \) & \( {\text{\small (76.6)}}_{\text{\large 39.8}} \) & \( {\text{\small (76.6)}}_{\text{\large 23.0}} \) & \({\text{\small (69.0)}}_{\text{\large 14.6}} \) & - 
        \\
        \cmidrule{2-8}
        & \textbf{\ALgname (Ours)}{$^\dagger$} & \HalfHatchedCircle \HalfHatchedCircle[0ex] 
        & \( {\text{\small (81.1)}}_{\text{\large \textbf{71.9}}} \) 
        & \( {\text{\small (77.0)}}_{\text{\large \textbf{60.1}}} \) 
        & \( {\text{\small (77.0)}}_{\text{\large \textbf{45.8}}} \) 
        & \( {\text{\small (66.2)}}_{\text{\large \textbf{39.4}}} \) 
        & \( {\text{\small (66.2)}}_{\text{\large \textbf{30.7}}} \) 
        \\
        \bottomrule
        \end{tabular}
        }
        \end{subcaptionblock}
\end{tabular}
\end{table}

\subsection{Main experiments}
\label{subsec:main_experiments}

\textbf{Results on CIFAR-10.}
In Table \ref{cifar-10-table}, we compare the performance of the baselines and \ALgname on CIFAR-10. Overall, \ALgname outperforms all existing state-of-the-art denoised smoothing (denoted by DS) approaches in every radii. For example, our method improves the best-performing denoised smoothing method \citep{carlinicertified} by $35.5\% \rightarrow 39.9\%$ at $\varepsilon = 1.00$. \ALgname also outperforms every randomzied smoothing techinque up to a radius of $\varepsilon \leq 1.00$. Even though our method slightly underperforms at higher radii in terms of certified accuracy, we note that \ALgname is the only denoised smoothing method which achieves a reasonable robustness at $\varepsilon > 1.00$. This means that our \ALgname effectively alleviates the distribution shift and hallucination issues observed in previous methods based on denoised smoothing \citep{carlinicertified}. We provide the detailed results in Appendix \ref{appendix:detail-results-main-experiments}.

\textbf{Results on ImageNet.}
In Table \ref{imagenet-table}, we compare the performance of the baselines and \ALgname on ImageNet, which is a far more complex dataset than CIFAR-10. In summary, \ALgname outperforms all existing state-of-the-art methods in every radii. In particular, our method surpasses the certified accuracy of diffusion denoised \citep{carlinicertified} by 9.9\% at $\varepsilon = 2.00$. 
In Table \ref{Imagenet-parameter-table}, we also compare 
{the architecture and trainable parameters of each method.} Our method even shows remarkable {parameter} efficiency, \ie we only update 0.9M parameters, which is 3\% of \citet{jeong2023confidence} and 0.2\% of \citet{jeong2024multi}. The overall results highlight the scalability of \ALgname, indicating its effectiveness in practical applications with only a small {parameter updates}. We provide the detailed results in Appendix \ref{appendix:detail-results-main-experiments} and further discuss the efficiency of LoRA \citep{hulora} on FT-CADIS in Appendix \ref{appendix:efficiency of LoRA}.

\begin{table}[t]
\caption{Comparison of the architectures and parameters between the previous state-of-the-art certified defense methods and \ALgname on ImageNet.}
\label{Imagenet-parameter-table}
\begin{center}
\resizebox{0.9\textwidth}{!}{%
\begin{tabular}{l|cccc}
\toprule

Method & \makecell{CAT-RS \\ \citep{jeong2023confidence}} & \makecell{Diffusion Denoised \\ \citep{carlinicertified}}
& \makecell{Multi-scale Denoised \\
\citep{jeong2024multi}} 
& \makecell{\textbf{\ALgname (Ours)}} \\ 
\midrule
Denoiser
& \makecell{-} 
& \makecell{Guided Diffusion \\ \citep{dhariwal2021diffusion}
} 
& \makecell{Guided Diffusion \\ \citep{dhariwal2021diffusion}
}  
& \makecell{Guided Diffusion \\ \citep{dhariwal2021diffusion}
} \\
\midrule
Classifier
& \makecell{ResNet-50 \\ \citep{he2016deep}} 
& \makecell{BEiT-large \\ \citep{baobeit} \\
} 
& \makecell{ViT-B/16 \\ \citep{dosovitskiy2020image}}     
& \makecell{ViT-B/16 (+LoRA) \\
\citep{dosovitskiy2020image}} \\ 
\midrule
Parameters     
& \makecell[l]{Denoiser : - \\ Classifier : 26M} 
& \makecell[l]{Denoiser : 552M \\ Classifier : 305M}
& \makecell[l]{Denoiser : 552M \\ Classifier : 87M}
& \makecell[l]{Denoiser : 552M \\ Classifier : 87M}\\
\midrule
Trainable        
& \makecell[l]{Denoiser : - \\ Classifier : 26M} 
& \makecell[l]{Denoiser : - \phantom{000} \\ Classifier : - \phantom{000}}
& \makecell[l]{Denoiser : 552M \\ Classifier : -}
& \makecell[l]{Denoiser : - \\ Classifier : \textbf{0.9M}}\\

\bottomrule
\end{tabular}
}
\end{center}
\end{table}

\begin{table}[t]
\centering
\caption{Comparison of ACR and certified accuracy for ablations of $\mathcal{L}^{\mathtt{\ALgname}}$ on CIFAR-10 with $\sigma = 1.00$.}
\label{ablation-loss-design}
\begin{tabular}{c}
    \centering
    \resizebox{0.9 \textwidth}{!}{%
    \begin{tabular}{lcccccccc}
    \toprule
        \multirow{2.5}{*}{Fine-tuning objective design} &
        \multirow{2.5}{*}{ACR} &
        \multicolumn{7}{c}{Certified Accuracy at $\varepsilon$ (\%)} \\ \cmidrule{3-9}
        && {0.00} & {0.25} & {0.50} & {0.75} & {1.00} & {1.25} & {1.50} \\
        \midrule
        $\mathcal{L}^{\mathtt{SCE}} + \lambda \cdot \mathcal{L}^{\mathtt{MAdv}}$ ($\mathcal{L}^{\mathtt{\ALgname}}$; \textbf{Ours}) & \textbf{0.784} & 48.1 & 43.5 & \textbf{40.6} & \textbf{36.9} & \textbf{32.5} & \textbf{28.6} & \textbf{23.7}  \\ \midrule
        \quad (w/o) Non-hallucinated condition of $\mathcal{L}^{\mathtt{SCE}}$ & 0.726 & 52.4 & 45.6 & 40.4 & 35.9 & 31.2 & 26.1 & 21.9 \\
        \quad (w/o) Mask of $\mathcal{L}^{\mathtt{MAdv}}$ & 0.374 & 11.2 & 10.9 & 10.4 & 10.2 & 10.2 & 10.2 & 10.2 \\ \midrule
        
        Cross-entropy loss $\mathcal{L}^{\mathtt{CE}}$ \citep{carlinicertified} & 0.633 & \textbf{54.4} & \textbf{45.8} & 39.3 & 33.2 & 28.1 & 22.4 & 17.3  \\
        
        \bottomrule
    \end{tabular}
    }   
\end{tabular}
\end{table}

\subsection{Ablation study}
\label{subsec:ablation}
In this section, we conduct an ablation study to further analyze the design of our proposed losses, the impact of updating the set of non-hallucinated images, and the component-wise effectiveness of our method. Unless otherwise stated, we report the test results based on a randomly sampled 1,000 images from the CIFAR-10 test set.

\textbf{Effect of overall loss design.}
Table \ref{ablation-loss-design} presents a comparison of variants of $\mathcal{L}^{\mathtt{\ALgname}}$, including: (a) removing the non-hallucinated condition of $\mathcal{L}^{\texttt{SCE}}$ in Eq. (\ref{eqn:ce_nh_loss}), (b) removing the masking condition of $\mathcal{L}^{\texttt{MAdv}}$ in Eq. (\ref{eqn:adv_nh_loss}), and (c) training with cross-entropy loss $\mathcal{L}^{\texttt{CE}}$ only. In summary, we observe that (a) using only non-hallucinated images for $\mathcal{L}^{\texttt{SCE}}$ achieves better ACR and effectively balances between accuracy and robustness. Additionally, we find that (b) the mask ``$\mathcal{D}_{\mathbf{x, nh}} = M$'' in $\mathcal{L}^{\texttt{MAdv}}$ is crucial for stable training, as it prevents the optimization of adversarial images that have lost the semantic of the original image; and (c) \ALgname demonstrates higher robustness and ACR by combining \LowLoss and \HighLoss.

\textbf{Effect of \HighLoss design.}
We further investigate the components of \HighLoss. Table \ref{appendix-design-of-high-loss} presents three variants of $\mathcal{L}^{\mathtt{MAdv}}$ in Eq. (\ref{eqn:adv_nh_loss}): (a) replacing the consistent target $\hat{y}$ with the assigned label $y$, (b) substituting the outer maximization with an average-case, and (c) combining both (a) and (b). Overall, we find that our proposed $\mathcal{L}^{\mathtt{MAdv}}$ demonstrates superior ACR compared to the variants, achieving the highest certified robustness while maintaining satisfactory clean accuracy. It shows that both design choices, \ie maximizing loss over adversarial images and using soft-labeled adversarial targets, are particularly effective.

\textbf{Effect of iterative update of $\mathcal{D}_{\mathbf{x, nh}}$.}
Our \ALgname iteratively updates the set of non-hallucinated images, \ie $\mathtt{denoise}(\mathbf{x}+\delta) \in \mathcal{D}_{\mathbf{x, nh}}$, to deal with the distribution shift from the pre-training distribution (clean images) to fine-tuning distribution (denoised images). Table \ref{ablation-classifier-state} shows the effect of $\mathcal{D}_{\mathbf{x, nh}}$ on varying $\sigma \in \{0.25, 0.50, 1.00\}$. For all noise levels, the iterative update strategy shows higher ACR with higher robustness. We find that the fine-tuning classifier increases the ratio of applying $\mathcal{L}^{\texttt{MAdv}}$ (see Figure~\ref{fig:effect_of_iterative_update} in Appendix~\ref{appendix:detail-effect-iterative-update}), \ie $f_{\texttt{clf}}$ gradually classifies all the denoised images of $\mathbf{x}$ correctly, thereby focusing on maximizing robustness and achieving a better trade-off between accuracy and robustness \citep{zhang2019theoretically}. 

\begin{figure}[t]
    \centering
    \begin{subcaptionblock}{0.49\linewidth}
        \centering
        \includegraphics[width=0.9\linewidth]{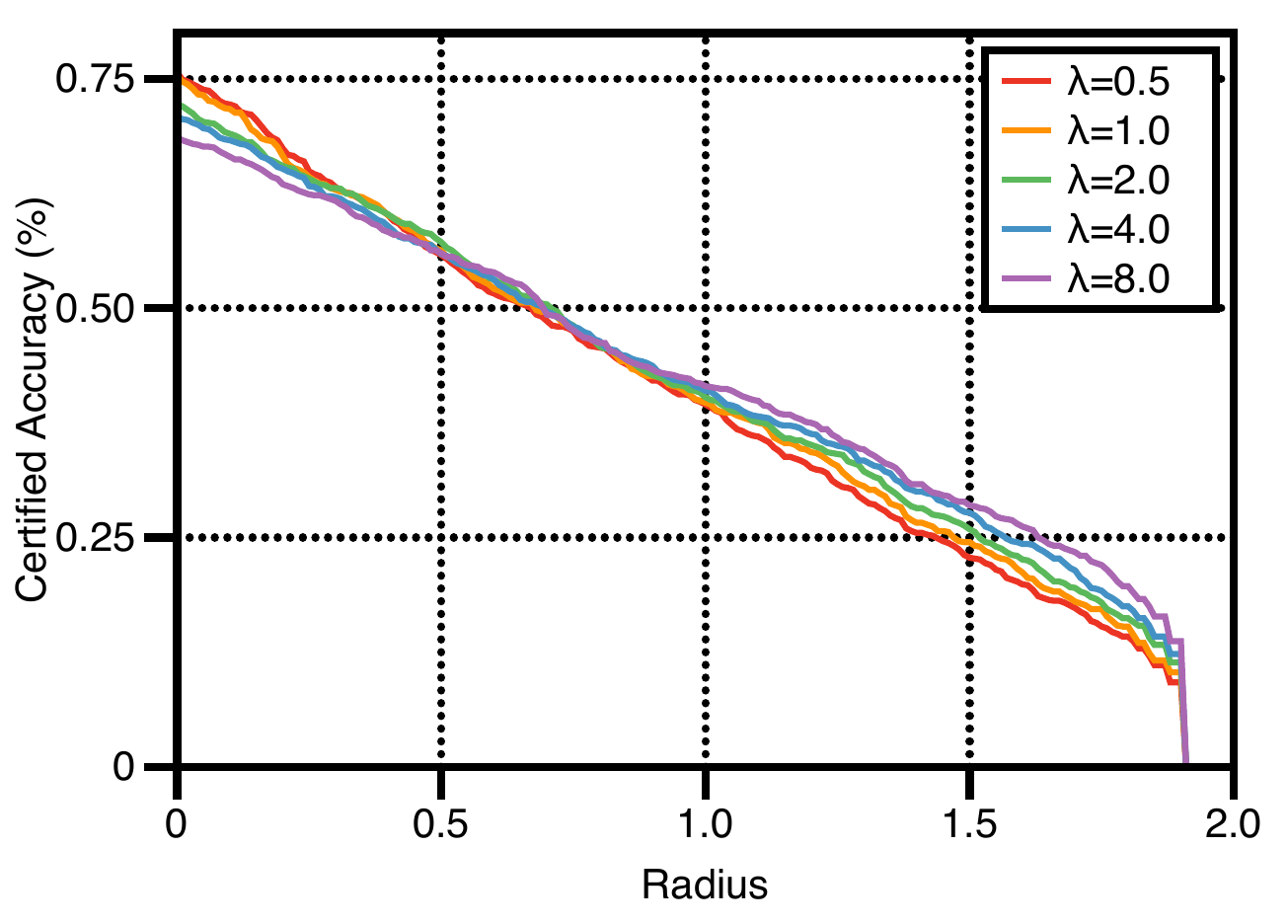}
        \caption{Effect of $\lambda$}
        \label{fig:ablation_lambda}
    \end{subcaptionblock}
    \vspace{1em} 
    \begin{subcaptionblock}{0.49\linewidth}
        \centering
    \includegraphics[width=0.9\linewidth]{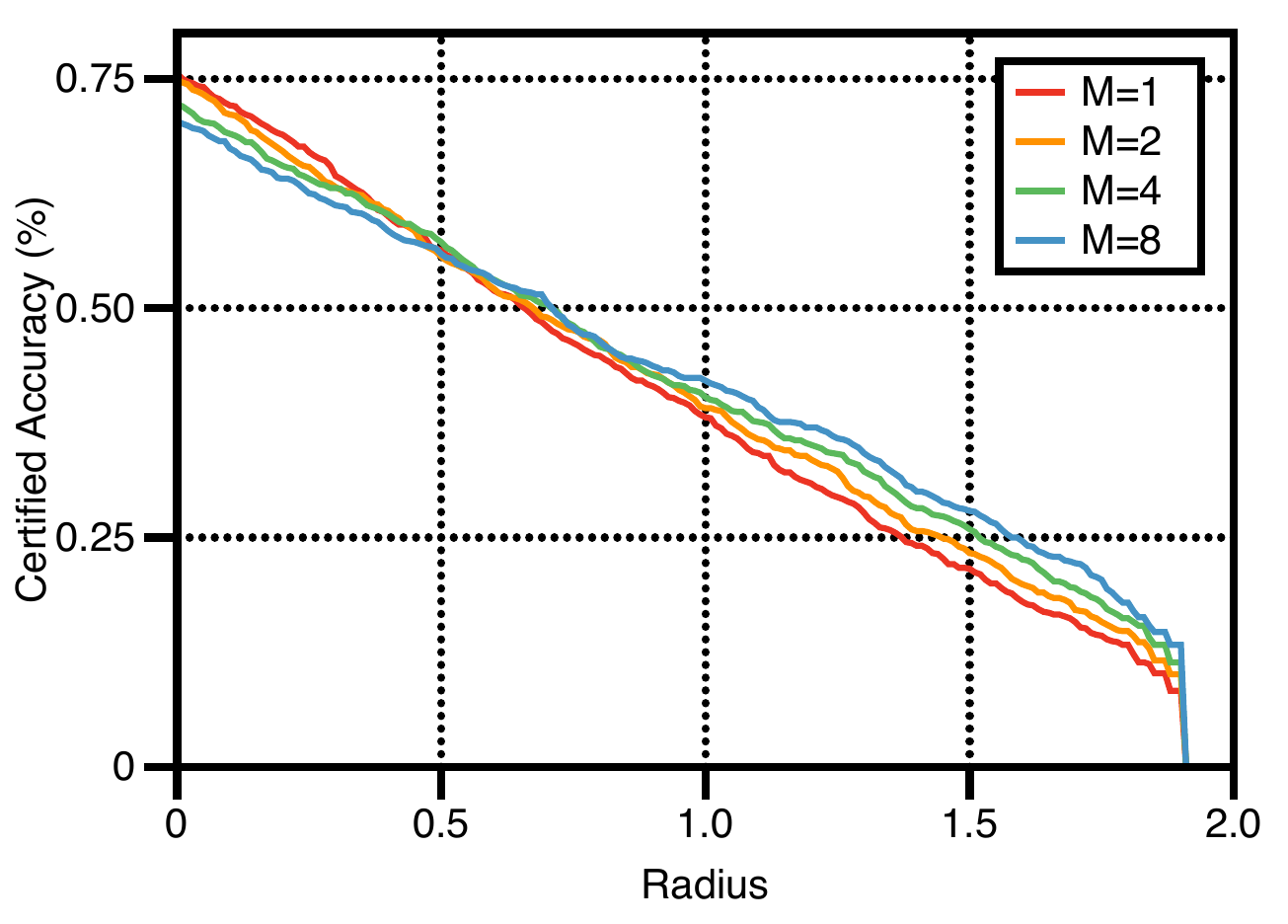}
        \caption{Effect of $M$}
        \label{fig:ablation_M}
    \end{subcaptionblock}
    \caption{Comparison of certified accuracy for components in \ALgname, (a) $\lambda$ and (b) $M$, on CIFAR-10. We plot the results at $\sigma=0.50$. We provide detailed results in Appendix \ref{appendix:detail-results-labmda-M}.}
    \label{fig:ablation_lambda_and_M}
\end{figure}

\textbf{Effect of $\lambda$.}
In the fine-tuning objective of \ALgname in Eq. (\ref{eqn:overall_objective}), \(\lambda\) determines the ratio between \(\mathcal{L}^{\mathtt{MAdv}}\) and \(\mathcal{L}^{\mathtt{SCE}}\). Figure \ref{fig:ablation_lambda} illustrates how \(\lambda\) affects the certified accuracy across different radii, with \(\lambda\) varying in \(\{0.5, 1.0, 2.0, 4.0, 8.0\}\) and \(\sigma = 0.50\). As \(\lambda\) increases, the robustness at high radii improves although the clean accuracy decreases, \ie the trade-off between clean accuracy and robustness.

\textbf{Effect of $M$.}
Figure \ref{fig:ablation_M} shows the impact of $M$ on the model when varying $M \in \{1, 2, 4, 8\}$. The robustness of the smoothed classifier improves as $M$ increases, while the clean accuracy decreases. With a higher $M$, the model is exposed to more denoised images included in $\mathcal{D}_{\mathbf{x, nh}}$, reducing the distribution shift from clean images to denoised images. This increases the confidence of the smoothed classifier, \ie the accuracy on denoised images, resulting in more robust predictions.

\begin{table}[t]
\centering
\caption{Comparison of ACR and certified accuracy for the ablation of the update of $\mathcal{D}_{\mathbf{x, nh}}$ on CIFAR-10.}
\label{ablation-classifier-state}
\begin{tabular}{c}
    \centering
    \resizebox{0.8\textwidth}{!}{%
    \begin{tabular}{cccccccccc}
    \toprule
        \multirow{2.5}{*}{{Noise}} & 
        \multirow{2.5}{*}{{Update of $\mathcal{D}_{\mathbf{x, nh}}$}} & 
        \multirow{2.5}{*}{{ACR}} &
        \multicolumn{7}{c}{Certified Accuracy at $\varepsilon$ (\%)} \\ \cmidrule{4-10}
        &&& {0.00} & {0.25} & {0.50} & {0.75} & {1.00} & {1.25} & {1.50} \\
        \toprule
        \multirow{2}{*}{$\sigma$ = 0.25} 
        & \textcolor{SJRed}{\xmark} & 0.632 & \textbf{91.1} & \textbf{80.4} & 66.7 & 49.0 & 0.0 & 0.0 & 0.0\\
        & \textcolor{SJViolet}{\cmark} & \textbf{0.642} & 87.9 & 78.7 & \textbf{68.0} & \textbf{54.0} & 0.0 & 0.0 & 0.0 \\
        
        \midrule
        \multirow{2}{*}{$\sigma$ = 0.50} 
        & \textcolor{SJRed}{\xmark} & 0.765 & \textbf{75.4} & \textbf{66.9} & 56.0 & 46.0 & 36.2 & 28.7 & 21.6 \\
        & \textcolor{SJViolet}{\cmark} & \textbf{0.806} & 72.2 & 64.1 & \textbf{57.2} & \textbf{48.1} & \textbf{40.3} & \textbf{34.1} & \textbf{25.9} \\
        
        \midrule
        \multirow{2}{*}{$\sigma$ = 1.00} 
        & \textcolor{SJRed}{\xmark} & 0.626 & \textbf{53.4} & \textbf{45.9} & 38.2 & 32.9 & 27.3 & 22.5 & 16.4 \\
        & \textcolor{SJViolet}{\cmark} & \textbf{0.783} & 48.1 & 43.5 & \textbf{40.6} & \textbf{36.9} & \textbf{32.4} & \textbf{28.5} & \textbf{23.8} \\
        
        \bottomrule
    \end{tabular}
    }   
\end{tabular}
\end{table}

\begin{table}[t]
\centering
\caption{Comparison of ACR and certified accuracy for ablations of $\mathcal{L}^{\mathtt{MAdv}}$ on CIFAR-10 with $\sigma = 0.50$. Every design adopts the $\mathds{1} {[{|\mathcal{D}_{\mathbf{x, nh}}|=M}]}$ masking condition.}
\label{appendix-design-of-high-loss}
\begin{tabular}{c}
    \centering
    \resizebox{0.9\textwidth}{!}{%
    \begin{tabular}{lcccccccc}
    \toprule
        \multirow{2.5}{*}{{Adversarial objective design}} & 
        \multirow{2.5}{*}{{ACR}} &
        \multicolumn{7}{c}{Certified Accuracy at $\varepsilon$ (\%)} \\ \cmidrule{3-9}
        && {0.00} & {0.25} & {0.50} & {0.75} & {1.00} & {1.25} & {1.50} \\
        \toprule

        (a) ${\mathrm{max}}_{i,\eta_i^*}\,\textnormal{KL}({f_{\texttt{clf}}}(\mathbf{x}+\eta_i^{*}),y)$ & 0.802 & 71.7 & 64.3 & 56.2 & 48.0 & 39.8 & 33.8 & 25.7 \\
        
        (b) $\frac{1}{M}\sum_{i}({\mathrm{max}}_{\eta_i^*}\,\textnormal{KL}({f_{\texttt{clf}}}(\mathbf{x}+\eta_i^{*}),\hat{y}))$ & 0.792 & \textbf{74.9} & \textbf{65.8} & 56.1 & 47.8 & 39.7 & 31.8 & 23.4\\
        
        (c) $\frac{1}{M}\sum_{i}
        ({\mathrm{max}}_{\eta_i^*}\,\textnormal{KL}({f_{\texttt{clf}}}(\mathbf{x}+\eta_i^{*}),y))$ & 0.792 & 74.8 & 64.9 & 57.0 & 48.0 & 39.9 & 31.5 & 23.0\\
        
        \midrule
        ${\mathrm{max}}_{i,\eta_i^*}\,\textnormal{KL}({f_{\texttt{clf}}}(\mathbf{x}+\eta_i^{*}),\hat{y}) \,(\mathcal{L}^{\mathtt{MAdv}}; \textbf{Ours})$ & \textbf{0.806} & 72.2 & 64.1 & \textbf{57.2} & \textbf{48.1} & \textbf{40.3} & \textbf{34.1} & \textbf{25.9} \\
        
        \bottomrule
    \end{tabular}
    }   
\end{tabular}
\end{table}

\section{Related Work}
\label{sec:related_work}
\vspace{-1em}
\textbf{Certified adversarial robustness.} Recently, various defenses have been proposed to build robust classifiers against adversarial attacks. In particular, \emph{certified defenses} have gained significant attention due to their guarantee of robustness \citep{wong2018provable, wang2018mixtrain, wang2018efficient, wong2018scaling}. 
Among them, \emph{randomized smoothing} \citep{lecuyer2019certified, li2019certified, cohen2019certified} shows the state-of-the-art performance by achieving the tight certified robustness guarantee over $\ell_2$-adversary \citep{cohen2019certified}. This approach converts any base classifier, \eg a neural network, into a provably robust smoothed classifier by taking a majority vote over random Gaussian noise. To maximize the robustness of the smoothed classifier, the base classifier should be trained with Gaussian-augmented images \citep{lecuyer2019certified, cohen2019certified, salman2019provably, zhaimacer, jeong2020consistency, jeong2023confidence}. For instance, \citet{salman2019provably} employed adversarial training \citep{madry2018towards} within the randomized smoothing framework, while \citet{jeong2020consistency} suggested training a classifier using simple consistency regularization. Moreover, \citet{jeong2023confidence} introduced sample-wise control of target robustness, motivated by the accuracy-robustness trade-off \citep{tsiprasrobustness, zhang2019theoretically} in smoothed classifiers. However, training base classifiers specifically for Gaussian-augmented images requires large training costs and thus these methods suffer from scalability issues in complex datasets, \eg the accuracy drops severely in the ImageNet dataset.

\textbf{Denoised smoothing.} Denoised smoothing alleviates the aforementioned scalability issue of randomized smoothing by introducing ``denoise-and-classify'' strategy. This approach allows randomized smoothing to be applied to any off-the-shelf classifier trained on clean images, \ie not specifically trained on Gaussian-augmented images, by adding a denoising step before feeding Gaussian-augmented images into the classifier. In recent years, {diffusion probabilistic models} have emerged as an ideal choice for the denoiser in the denoised smoothing scheme. In particular, \citet{lee2021provable} have initially explored the applicability of diffusion models in denoised smoothing, and \citet{carlinicertified} further observe that combining the latest diffusion models with an off-the-shelf classifier provides a state-of-the-art design for certified robustness. Meanwhile, \citet{jeong2024multi} investigate the trade-off between robustness and accuracy in denoised smoothing, and proposed a multi-scale smoothing scheme that incorporates denoiser fine-tuning.

Our work aims to improve the certified robustness of smoothed classifiers in denoised smoothing, which is determined by the average accuracy of the off-the-shelf classifiers under denoised images. We improve such robustness by addressing hallucination and distribution shift issues of denoised images. Specifically, we focus on filtering out hallucinated images based on the confidence of off-the-shelf classifiers, and then fine-tune off-the-shelf classifiers with non-hallucinated images.

\section{Conclusion}
\label{sec:conclusion}

We propose \ALgname, a scalable fine-tuning strategy of off-the-shelf classifiers for certified robustness. Specifically, we propose to use the \emph{confidence} of off-the-shelf classifiers to mitigate the intrinsic drawbacks of the denoised smoothing framework, \ie hallucination and distribution shift. We also demonstrate that this can be achieved by updating only 1\% of the total parameters. We hope that our method could be a meaningful step for the future research to develop a scalable approach for certified robustness.

{\bf Limitation and future work.}
In this work, we apply an efficient training technique for off-the-shelf classifiers based on LoRA \citep{hulora}. Nevertheless, certification remains a bottleneck for throughput, due to its majority voting process involving a large number of forward inferences, \ie $N=100,000$. An important future work would be to accelerate the certification process for a more practical deployment of our method. In addition, certain public vision APIs do not allow us to access the underlying off-the-shelf classifiers, \emph{i.e.}, black-box. In such cases, our method is not directly applicable, and further research on training methods that are independent of model parameters, such as prompt-tuning \citep{jia2022visual}, will be necessary.

\section*{Acknowledgements}
This work was conducted by Center for Applied Research in Artificial Intelligence (CARAI) grant funded by Defense Acquisition Program Administration (DAPA) and Agency for Defense Development (ADD) (UD230017TD), and supported by Institute for Information \& communications Technology Promotion (IITP) grant funded by the Korea government (MSIT) (No.RS-2019-II190075, Artificial Intelligence Graduate School Program (KAIST); No. RS-2019-II190079, Artificial Intelligence Graduate School Program (Korea University)), and by Culture, Sports and Tourism R\&D Program through the Korea Creative Content Agency grant funded by the Ministry of Culture, Sports and Tourism in 2024 (Project Name: International Collaborative Research and Global Talent Development for the Development of Copyright Management and Protection Technologies for Generative AI, Project Number: RS-2024-00345025).

\bibliographystyle{tmlr}

\clearpage
\appendix
\begin{center}
{\bf {\LARGE Supplementary Material}} \\
\vspace{0.05in}
{\bf {Appendix: Confidence-aware Denoised Fine-tuning of \\ Off-the-shelf Models for Certified Robustness}}
\end{center}

\section{Training Procedure of \ALgname}
\label{appendix:training Procedure}

\begin{algorithm}
\caption{\Algname (\ALgname)}
\label{appendix:alg:FT-CADIS}
\begin{algorithmic}[1]
\Require training sample $(\mathbf{x},y)$. variance of Gaussian noise $\sigma$. number of noises $M$. off-the-shelf classifier $f_{\texttt{clf}}$. attack \(\ell_2\)-norm $\varepsilon > 0$. adversarial target $\hat{y} \in \Delta^{K-1}$. coefficient of \HighLoss $\lambda > 0$.
\algrule
\State $\text{Generate} \,\, \mathbf{\hat{x}}_1 = \Call{NoiseAndDenoise}{\mathbf{x}_1, \sigma}, \,\, \cdots, \,\, \mathbf{\hat{x}}_M = \Call{NoiseAndDenoise}{\mathbf{x}_M, \sigma}$ \Comment{$\mathbf{x}_i$: copy of $\mathbf{x}$}
\State Identify $\mathcal{D}_{\mathbf{x,nh}} = \{\mathbf{\hat{x}}_i \mid f_{\texttt{clf}}(\mathbf{\hat{x}}_i) = y, \,i \in [1,...,M]\}$
\For{$i=1$ to $M$}
    \State $\mathcal{L}_i \gets \mathbb{CE}({f_{\texttt{clf}}}(\mathbf{\hat{x}}_i),y)$
    \State $\eta_i^{*} \gets \underset{\lVert \eta_i^{*} - \eta_{i} \rVert_2 \leq \varepsilon}{\mathrm{arg\,max}} \,\, \textnormal{KL}({f_{\texttt{clf}}}(\mathbf{x}+\eta_i^{*}),\hat{y}), \,\, \eta_i := \,\, \mathbf{\hat{x}}_i - \mathbf{x}$
\EndFor
\State $\mathcal{L}_{1:M}^\pi, \textsc{indices} \gets \mathtt{argsort}(\mathcal{L}_{1:M}), \,\, \mathcal{D}_{\mathbf{x, nh}}^\pi \gets \{\mathbf{\hat{x}}^{\pi}_{\textsc{indices}.\texttt{index}(i)} \mid \mathbf{\hat{x}}_i \in \mathcal{D}_{\mathbf{x, nh}}\}$
\If{$\mathcal{D}_{\mathbf{x, nh}}^\pi \neq \emptyset$}
    \State $\mathcal{L}^{\mathtt{SCE}} \gets \frac{1}{M}(\sum_{\mathbf{\hat{x}}^{\pi}_{i} \in \mathcal{D}_{\mathbf{x,nh}}^\pi} \mathcal{L}_i^\pi)$
\Else
    \State $\mathcal{L}^{\mathtt{SCE}} \gets \frac{1}{M}(\mathcal{L}_1^\pi)$ \Comment{$\mathcal{L}_1^\pi$: lowest cross-entropy loss}
\EndIf
\State $\mathcal{L}^{\mathtt{MAdv}} \gets \mathds{1}[|\mathcal{D}_{\mathbf{x, nh}}|=M] \cdot \underset{i}{\mathrm{max}} \,\, \textnormal{KL}({f_{\texttt{clf}}}(\mathbf{x}+\eta_i^{*}), \,\, \hat{y})$
\State $\mathcal{L}^{\mathtt{\ALgname}} \gets \mathcal{L}^{\mathtt{SCE}} + \lambda \cdot \mathcal{L}^{\mathtt{MAdv}}$
\end{algorithmic}
\end{algorithm}

\begin{algorithm}
\caption{Noise-and-Denoise Procedure \citep{carlinicertified}}
\label{appendix:alg:noising-and-denoising}
\begin{algorithmic}[1]
\Function{NoiseAndDenoise}{$\mathbf{x}, \sigma$}:
    \State $t^*, \alpha_{t^*} \gets \Call{GetTimestep}{\sigma}$
    \State $\mathbf{x}_{t^*} \gets \sqrt{\alpha_{t^*}}(\mathbf{x} + \delta), \,\, \delta \sim \mathcal{N}(0, \sigma^2 \mI)$
    \State $\mathbf{\hat{x}} \gets \mathtt{denoise}(\mathbf{x}_{t^*}; t^*)$ \Comment{$\mathtt{denoise}$ : one-shot diffusion denoising process}
    \State \Return $\hat{\mathbf{x}}$
\EndFunction
\State

\Function{GetTimestep}{$\sigma$}:
    \State $t^* \gets \text{find the timestep } t \text{ s.t. } \sigma^2 = \frac{1 - \alpha_t}{\alpha_t}$ \Comment{$\alpha_t$ : noise level constant of diffusion model}
    \State \Return $t^*, \alpha_{t^*}$
\EndFunction

\end{algorithmic}
\end{algorithm}

\clearpage
\section{Experimental Details}
\label{appendix:experimental details}
\subsection{Datasets}
\label{appendix:Datasets}

\textbf{CIFAR-10} \citep{alex2009learning} consists of 60,000 RGB images of size 32$\times$32, with 50,000 images for training and 10,000 for testing. Each image is labeled as one of 10 classes. We apply the standard data augmentation, including random horizontal flip and random translation up to 4 pixels, as used in previous works \citep{cohen2019certified, salman2019provably, zhaimacer, jeong2020consistency,jeong2021smoothmix,jeong2023confidence}. No additional normalization is applied except for scaling the pixel values from [0,255] to [0.0, 1.0] when converting image into a tensor. The full dataset can be downloaded at \href{https://www.cs.toronto.edu/~kriz/cifar.html}{https://www.cs.toronto.edu/~kriz/cifar.html}.

\textbf{ImageNet} \citep{ILSVRC15} consists of 1.28 million training images and 50,000 validation images, each labeled into one of 1,000 classes. For the training images, we apply 224$\times$224 randomly resized cropping and horizontal flipping. For the test images, we resize them to 256$\times$256 resolution, followed by center cropping to 224$\times$224. Similar to CIFAR-10, no additional normalization is applied except for scaling the pixel values to [0.0, 1.0]. The full dataset can be downloaded at \href{https://image-net.org/download}{https://image-net.org/download}.

\subsection{Training}
\label{appendix:Training}

\textbf{Noise-and-Denoise Procedure.} We follow the protocol of \citet{carlinicertified} to obtain the denoised images for fine-tuning. Firstly, the given image $\mathbf{x}$ is clipped to the range [-1,1] as expected by the off-the-shelf diffusion models. Then, the perturbed  image is obtained from a certain diffusion time step according to the target noise level. Finally, we adopt a one-shot denoising, \ie outputting the best estimate for the denoised image in a single step, resulting in a denoised image within the range of [-1,1]. Since this range differs from the typical range of $[0, 1]$ assumed in prior works,  we set the target noise level to twice the usual level for training and certification. A detailed implementation can be found at \href{https://github.com/ethz-spylab/diffusion_denoised_smoothing} {https://github.com/ethz-spylab/diffusion-denoised-smoothing} and the algorithm is provided in Algorithm \ref{appendix:alg:noising-and-denoising}.

\textbf{CIFAR-10 fine-tuning.}
We conduct an end-to-end fine-tuning of a pre-trained ViT-B/16 \citep{dosovitskiy2020image}, considering different scenarios of $\sigma \in \{0.25, 0.50, 1.00\}$ for randomized smoothing. The same $\sigma$ is applied to both the training and certification. As part of the data pre-processing, we interpolate the dataset to 224$\times$224. Our fine-tuning follows the common practice of supervised ViT training. The default setting is shown in Table \ref{appendix:cifar10-finetuning}. We use the linear \emph{lr} scaling rule \citep{goyal2017accurate}: $\emph{lr} = \emph{base lr}\,\times\,\emph{batch size}\,\div\,256$. The batch size is calculated as $\emph{batch per GPU}\times\emph{number of GPUs}\,\times\,\emph{accum iter}\,\,\texttt{//}\, \emph{number of noises}$, where \emph{accum iter} denotes the batch accumulation hyperparameter.

\textbf{ImageNet fine-tuning.} We adopt LoRA \citep{hulora} to fine-tune a pre-trained ViT-B/16 \citep{dosovitskiy2020image} in a parameter-efficient manner. We use the same training scenarios as for CIFAR-10. As part of the data pre-processing, we interpolate the dataset to 384$\times$384. The default setting is shown in Table \ref{appendix:imagenet-finetuning}. Compared to end-to-end fine-tuning, we reduce the regularization setup, \eg weight decay, \emph{lr} decay, drop path, and gradient clipping. For LoRA fine-tuning, we freeze the original model except for the classification layer. Then, LoRA weights are incorporated into each query and value projection matrix of the self-attention layers of ViT. For these low-rank matrices, we use Kaiming-uniform initialization for weight $\mA$ and zeros for weight $\mB$, following the \href{https://github.com/microsoft/LoRA}{official code}. To implement LoRA with ViT, we refer to \href{https://github.com/JamesQFreeman/LoRA-ViT}{https://github.com/JamesQFreeman/LoRA-ViT}.

\subsection{Hyperparameters}
\label{appendix:Hyperparameters}

\begin{table}[htb!]
\centering
\caption{Denoised fine-tuning settings for the off-the-shelf classifier on CIFAR-10 and ImageNet.}
\label{appendix:finetuning-setting}
\begin{subcaptionblock}{\textwidth}
    \caption{CIFAR-10 end-to-end fine-tuning}
    \label{appendix:cifar10-finetuning}
    \centering
    \resizebox{0.7\textwidth}{!}{%
        \begin{tabular}{l|l}
        \toprule
            Configuration & Value \\
        \midrule
            Optimizer & AdamW \citep{loshchilovdecoupled}\\
            Optimizer momentum &  $\beta_1$, $\beta_2$ = 0.9, 0.999 \\
            Base learning rate &  5e-4 ($\sigma=0.25, 0.50$), 1e-4 ($\sigma = 1.00$)\\
            Weight decay & start, end = 0.04, 0.4 (cosine schedule)\\
            Layer-wise lr decay \citep {clark2020electra, baobeit} & 0.65 \\
            Batch size &  128 \\
            Learning rate schedule & cosine decay \citep{loshchilov2022sgdr} \\
            Warmup epochs \citep{goyal2017accurate} & 3 \\
            Training epochs & 30 (early stopping at 20)\\
            Drop path \citep{huang2016deep} & 0.2 \\
            Gradient clipping \citep{zhang2019gradient} & 0.3 \\
        \bottomrule
        \end{tabular}
    }
\end{subcaptionblock}

\vspace{0.5cm}

\begin{subcaptionblock}{\textwidth}
    \caption{ImageNet LoRA \citep{hulora} fine-tuning}
    \label{appendix:imagenet-finetuning}
    \centering
    \resizebox{0.7\textwidth}{!}{%
        \begin{tabular}{l|l}
        \toprule
            Configuration & Value \\
        \midrule
            Optimizer & AdamW \citep{loshchilovdecoupled}\\
            Optimizer momentum &  $\beta_1$, $\beta_2$ = 0.9, 0.999 \\
            Base learning rate &  2e-4 ($\sigma=0.25$), 4e-4 ($\sigma = 0.50, 1.00$)\\
            \multirow{2}{*}{Weight decay} 
            & start, end = 0.02, 0.2 ($\sigma=0.25$) \\ 
            & start, end = 0.01, 0.1 ($\sigma=0.50, 1.00$) \\
            Layer-wise lr decay \citep {clark2020electra, baobeit} & 0.8 ($\sigma=0.25$), 0.9 ($\sigma=0.50, 1.00$) \\
            Batch size &  128 \\
            Learning rate schedule & cosine decay \citep{loshchilov2022sgdr} \\
            Warmup epochs \citep{goyal2017accurate} & 1 \\
            Training epochs & 10 (early stopping at 5)\\
            Drop path \citep{huang2016deep} & 0.0 \\
            Gradient clipping \citep{zhang2019gradient} & 1.0 \\
        \midrule
            LoRA rank r & 4 \\
            LoRA scaler $\alpha$ & 4 \\
        \bottomrule
        \end{tabular}
    }
\end{subcaptionblock}

\end{table}

In our proposed loss functions (see Eqs. (\ref{eqn:ce_nh_loss}), (\ref{eqn:adv_nh_loss}), and (\ref{eqn:overall_objective})), there are two main hyperparameters: the coefficient \(\lambda\) for the \HighLoss, and the attack radius \(\varepsilon\) of \HighLoss. We have determined the optimal configurations for two hyperparameters through a simple grid search on $\lambda$ over [1,2,4] and $\varepsilon$ over [0.125, 0.25, 0.5, 1.0].

For CIFAR-10, we use \(\lambda = 1.0, 2.0, 4.0\) for \(\sigma = 0.25, 0.50, 1.00\), respectively. Assuming that \(\texttt{denoise}(\mathbf{x} + \delta) \approx \mathbf{x}\) with high probability, we adopt a small \(\varepsilon = 0.25\) by default, which is increased to 0.50 after 10 epochs only for \(\sigma = 1.00\). For ImageNet, we use \(\lambda = 2.0, 1.0, 2.0\) for \(\sigma = 0.25, 0.50, 1.00\) respectively, and \(\varepsilon\) is fixed at 0.25 for all noise levels. Although the number of noises \(M\) and the number of attack steps \(T\) can also be tuned for better performance, we fix \(M = 4\) and \(T = 4\) for CIFAR-10. For ImageNet, we fix \(M = 2\) and \(T = 1\) to reduce the overall training cost. Additional training configurations are provided in Table \ref{appendix:finetuning-setting}. Due to the extensive training cost of large models, we have adjusted some training configurations for the ablation study, \eg the warmup and training epochs are reduced to 2 and 20, with $\varepsilon$ doubled after 10 epochs.

\subsection{Computing infrastructure}
\label{appendix:Computing infrastructure}
In summary, we conduct our experiments using NVIDIA GeForce RTX 2080 Ti GPUs for CIFAR-10, NVIDIA GeForce RTX 3090 and NVIDIA RTX A6000 GPUs for ImageNet. In the CIFAR-10 experiments, we utilize 4 NVIDIA GeForce RTX 2080 Ti GPUs for fine-tuning per run, resulting in $\sim$8 hours of training cost. During the certification, we use 7 NVIDIA GeForce RTX 2080 Ti GPUs for data splitting, taking $\sim$9 minutes per image (with $N = 100,000$ for each inference) to perform a single pass of smoothed inference. In the ImageNet experiments, we utilize 4 NVIDIA RTX A6000 GPUs for fine-tuning per run, observing $\sim$51 hours of training cost. During the certification, 8 NVIDIA GeForce RTX 3090 GPUs are used in parallel, taking $\sim$4 minutes per image (with $N = 10,000$ for each inference) to complete a single pass of smoothed inference.

\clearpage
\subsection{Detailed results on main experiments}
\label{appendix:detail-results-main-experiments}

\begin{table}[htb!]
\centering
\caption{Certified accuracy of \ALgname for varying levels of Gaussian noise $\sigma$ on CIFAR-10 and ImageNet. Values in bold-faced indicate the ones reported in Table \ref{cifar-10-table} for CIFAR-10 and Table \ref{imagenet-table} for ImageNet.}
\label{appendix-main-experiments}
\begin{tabular}{cc}
    \begin{subcaptionblock}{0.6\textwidth}
        \caption{CIFAR-10}
        \centering
        \resizebox{\textwidth}{!}{%
            \begin{tabular}{cccccccc}
            \toprule
                \multirow{2.5}{*}{{Noise}} & \multicolumn{7}{c}{Certified Accuracy at $\varepsilon$ (\%)} \\ \cmidrule{2-8}
                & 0.00 & 0.25 & 0.50 & 0.75 & 1.00 & 1.25 & 1.50 \\
            \toprule
                $\sigma$ = 0.25 & 88.7 & \textbf{80.3} & \textbf{68.4} & \textbf{54.5} & 0.0 & 0.0 & 0.0 \\
                $\sigma$ = 0.50 & 74.9 & 67.3 & 58.7 & 49.2 & \textbf{39.9} & \textbf{31.6} & \textbf{23.5} \\
                $\sigma$ = 1.00 & 49.6 & 45.5 & 41.0 & 36.8 & 32.5 & 28.4 & 24.2 \\
            \bottomrule
            \end{tabular}
        }
    \end{subcaptionblock}
    \vspace{1em}
    \\
    \begin{subcaptionblock}{0.55\textwidth}
        \caption{ImageNet}
        \centering
        \resizebox{\textwidth}{!}{%
            \begin{tabular}{ccccccc}
            \toprule
                \multirow{2.5}{*}{{Noise}} & \multicolumn{6}{c}{Certified Accuracy at $\varepsilon$ (\%)} \\ \cmidrule{2-7}
                & 0.00 & 0.50 & 1.00 & 1.50 & 2.00 & 2.50 \\
            \toprule
                $\sigma$ = 0.25 & 81.1 & \textbf{71.9} & 0.0 & 0.0 & 0.0 & 0.0 \\
                $\sigma$ = 0.50 & 77.0 & 69.3 & \textbf{60.1} & \textbf{45.8} & 0.0 & 0.0 \\
                $\sigma$ = 1.00 & 66.2 & 60.7 & 54.0 & 46.4 & \textbf{39.4} & \textbf{30.7} \\
            \bottomrule
            \end{tabular}
        }
    \end{subcaptionblock}
\end{tabular}
\end{table}

\vspace{1cm}

\begin{figure}[htb!]
    \centering
    \begin{subcaptionblock}{0.45\linewidth}
        \centering
        \includegraphics[width=\linewidth]{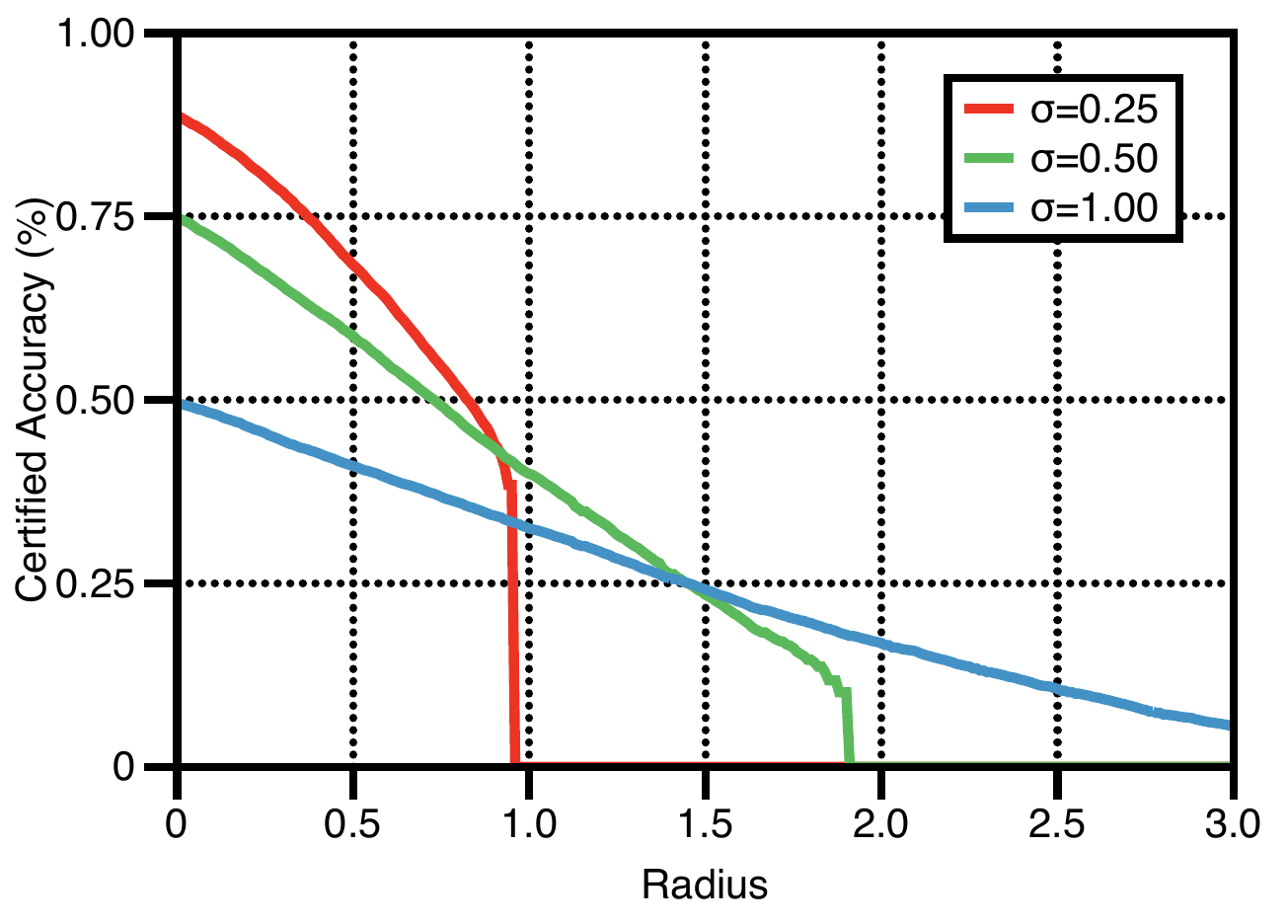}
        \caption{CIFAR-10}
        \label{fig:cifar10_certified_accuracy_figure}
    \end{subcaptionblock}
    \hfill
    \begin{subcaptionblock}{0.45\linewidth}
        \centering
        \includegraphics[width=\linewidth]{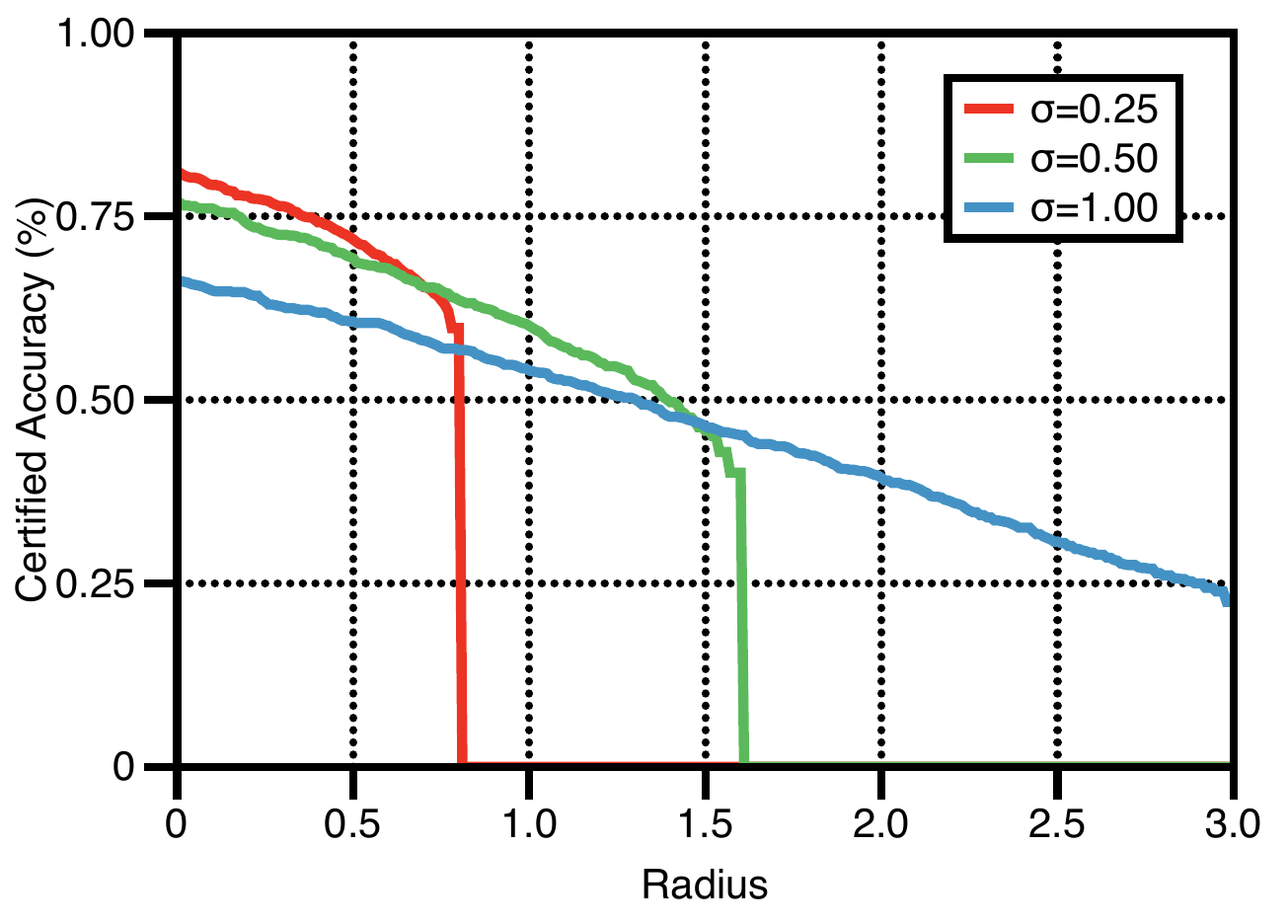}
        \caption{ImageNet}
        \label{fig:imagenet_certified_accuracy_figure}
    \end{subcaptionblock}
    \caption{Certified accuracy of \ALgname at different levels of Gaussian noise $\sigma \in \{0.25, 0.50, 1.00\}$. Upper bounds in radius are calculated with $N = 100{,}000$ for CIFAR-10 and $N = 10{,}000$ for ImageNet.}
    \label{fig:certified_accuracy_figure}
\end{figure}

\clearpage
\section{Detailed Results on Effect of \texorpdfstring{$\lambda$}{Lg} and \texorpdfstring{$M$}{Lg}}
\label{appendix:detail-results-labmda-M}

\begin{table}[htb!]
\centering
\caption{Comparison of ACR and certified  accuracy for ablations of varying $\lambda$ on CIFAR-10 with $\sigma$ = 0.50.}
\label{appendix-ablation-lambda}
\begin{tabular}{c}
    \centering
    \resizebox{0.7\textwidth}{!}{%
    \begin{tabular}{ccccccccc}
    \toprule
        \multirow{2.5}{*}{{Setups}} & 
        \multirow{2.5}{*}{{ACR}} &
        \multicolumn{7}{c}{Certified Accuracy at $\varepsilon$ (\%)} \\ \cmidrule{3-9}
        && {0.00} & {0.25} & {0.50} & {0.75} & {1.00} & {1.25} & {1.50} \\
        \toprule
        $\lambda$ = 0.50 & 0.786 & \textbf{75.7} & \textbf{64.9} & 55.8 & 47.5 & 39.5 & 30.8 & 22.8 \\
        $\lambda$ = 1.00 & 0.797 & 75.3 & 64.3 & 56.3 & 47.9 & 39.7 & 32.8 & 24.5 \\
        $\lambda$ = 2.00 & 0.806 & 72.2 & 64.1 & \textbf{57.2} & \textbf{48.1} & 40.3 & 34.1 & 25.9 \\
        $\lambda$ = 4.00 & 0.814 & 70.9 & 63.3 & 55.9 & 48.0 & 41.0 & 35.0 & 27.7 \\
        $\lambda$ = 8.00 & \textbf{0.823} & 68.6 & 62.4 & 56.0 & 47.6 & \textbf{41.5} & \textbf{35.9} & \textbf{28.5}\\
        \bottomrule
    \end{tabular}
    }   
\end{tabular}
\end{table}
\begin{table}[htb!]
\centering
\caption{Comparison of ACR and certified accuracy for ablations of varying $M$ on CIFAR-10 with $\sigma$ = 0.50.}
\label{appendix-ablation-M}
\begin{tabular}{c}
    \centering
    \resizebox{0.7\textwidth}{!}{%
    \begin{tabular}{ccccccccc}
    \toprule
        \multirow{2.5}{*}{{Setups}} & 
        \multirow{2.5}{*}{{ACR}} &
        \multicolumn{7}{c}{Certified Accuracy at $\varepsilon$ (\%)} \\ \cmidrule{3-9}
        && {0.00} & {0.25} & {0.50} & {0.75} & {1.00} & {1.25} & {1.50} \\
        \toprule
        $M$ = 1 & 0.773 & \textbf{75.6} & \textbf{67.0} & 56.3 & 46.2 & 38.1 & 29.4 & 21.6\\
        $M$ = 2 & 0.790 & 74.9 & 65.4 & 55.6 & 47.6 & 39.1 & 32.2 & 23.3 \\
        $M$ = 4 & 0.806 & 72.2 & 64.1 & \textbf{57.2} & \textbf{48.1} & 40.3 & 34.1 & 25.9 \\
        $M$ = 8 & \textbf{0.817} & 70.4 & 62.5 & 55.9 & 47.9 & \textbf{42.1} & \textbf{35.8} & \textbf{27.9} \\
        \bottomrule
    \end{tabular}
    }   
\end{tabular}
\end{table}

\section{Effect of Iterative Update of \texorpdfstring{$\mathcal{D}_{\mathbf{x, nh}}$}{Lg} on \texorpdfstring{$\mathcal{L}^{\texttt{MAdv}}$}{Lg}}
\label{appendix:detail-effect-iterative-update}

\begin{figure}[htb!]
    \centering
    \begin{subcaptionblock}{0.60\linewidth}
        \centering
        \includegraphics[width=\linewidth]{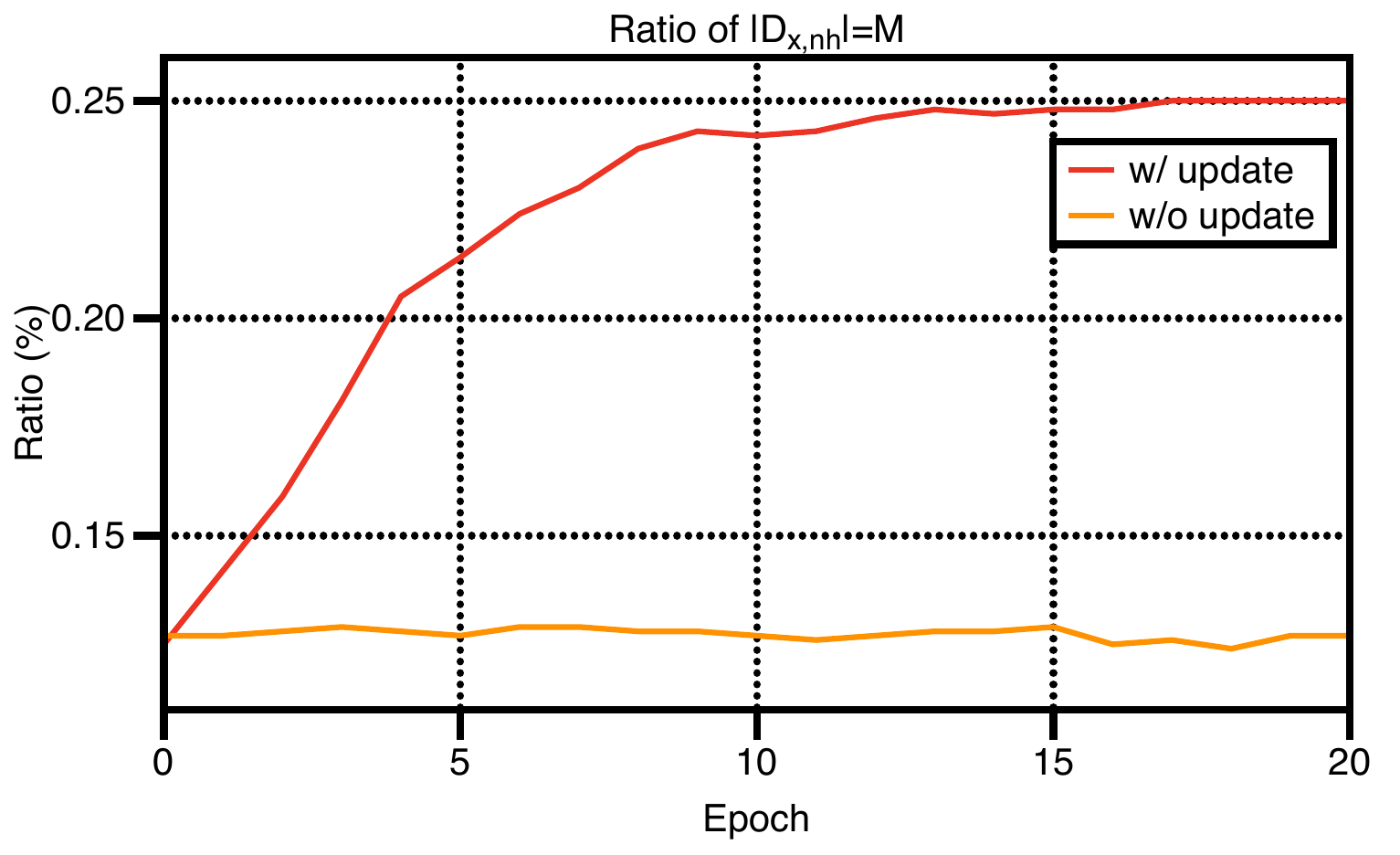}
    \end{subcaptionblock}
    \caption{Change in the ratio of $|\mathcal{D}_{\mathbf{x, nh}}| = M$, \ie ratio of clean images $\mathbf{x}$ satisfying the masking condition of $\mathcal{L}^{\mathtt{MAdv}}$, during fine-tuning on CIFAR-10 with $\sigma=1.00$, depending on whether $\mathcal{D}_{\mathbf{x, nh}}$ is being updated or not. In the legend, red indicates that $\mathcal{D}_{\mathbf{x, nh}}$ is iteratively updated, while orange indicates that $\mathcal{D}_{\mathbf{x, nh}}$ is fixed.}
    \label{fig:effect_of_iterative_update}
\end{figure}

\clearpage
\section{Details on Certifying Robustness of \ALgname}
\label{appendix:explanation-certification}

We simply follow the common evaluation framework of the baselines \citep{carlinicertified, jeong2024multi}. In denoised smoothing framework \citep{salman2020denoised}, the robustness of the smoothed classifier $g$ is guaranteed based on the accuracy of the off-the-shelf classifier $f_\texttt{clf}$ under denoised images, \viz $\mathbb{P}_{\delta}[f_{\texttt{clf}}(\texttt{denoise}(\mathbf{x} + \delta))=y] =: p_{g}(\mathbf{x},y) $. However, since $f_{\texttt{clf}}$ is a high-dimensional neural network, this accuracy cannot be computed directly. Instead, we estimate it using the practical Monte-Carlo based algorithm \textsc{Certify} from \citet{cohen2019certified}.

\textsc{Certify} algorithm consists of two main procedures: (a) given input \textbf{x}, identifying the most probable output class $\hat{c}_A$ of $f$, and (b) computing a lower-bound on the probability that output of $f$ is $\hat{c}_A$:

\begin{itemize}
    \item[(a)] Using a small number $n_0$ (\eg $n_0$ = 100) of denoised images and taking a majority vote over the outputs of $f_{\texttt{clf}}$, \ie computing   $f_{\texttt{clf}}(\texttt{denoise}(\mathbf{x} + \delta))$ $n_0$ times to identify the most frequent class $\hat{c}_A$.
    
    \item[(b)] Using a large number $n$ (\eg $n$ = 100,000) of denoised images to estimate the lower bound of $\mathbb{P}_{\delta}[f_{\texttt{clf}}(\texttt{denoise}(\mathbf{x} + \delta))= \hat{c}_A]$, \viz $\underline{p_{g}}(\mathbf{x},\hat{c}_A)$, considering the significance level $\alpha$ (see \citet{cohen2019certified} for details).
\end{itemize}

Finally, the certified $\ell_{2}$-norm radius of smoothed classifier, derived by \citet{cohen2019certified}, is calculated as $\sigma \cdot \Phi^{-1}(\underline{p_{g}}(\mathbf{x},\hat{c}_A))$ when $\underline{p_{g}}(\mathbf{x},\hat{c}_A)) > \frac{1}{2}$; otherwise return \textsc{Abstain}. It verifies that the smoothed classifier $g$ does not change its output within an $\ell_{2}$ ball of radius $\sigma \cdot \Phi^{-1}(\underline{p_{g}}(\mathbf{x},\hat{c}_A))$ around any input $\mathbf{x}$.

\clearpage
\section{Analysis on Efficiency of FT-CADIS with LoRA}
\label{appendix:efficiency of LoRA}

\begin{table}[htb!]
\centering
\caption{Comparison of fine-tuning costs and ACR between the baselines \citep{carlinicertified, jeong2024multi} and FT-CADIS on ImageNet with $\sigma=0.50$. $\dagger$ indicates that the numbers are taken from the original paper.}
\label{appendix-training-cost}
\resizebox{0.9\textwidth}{!}{%
\begin{tabular}{lccccc}
\toprule
Method & \makecell{LoRA \\ \citep{hulora}} & ACR & \makecell{Fine-tuned model} & 
\makecell{Trainable \\ parameters} & GPU days \\ \midrule
\makecell[l]{Diffusion Denoised \\ \citep{carlinicertified}} & - & 0.896 & - & 0M & - \\ \midrule
\makecell[l]{Multi-scale Denoised{$^\dagger$} \\ \citep{jeong2024multi}} & \textcolor{SJRed}{\xmark} & 0.743 & \makecell{Guided Diffusion \\ \citep{dhariwal2021diffusion}} & 552M & 32 \\ 
\midrule
\multirow{3.5}{*}{\makecell[l]{\textbf{FT-CADIS (Ours)}}} & \textcolor{SJRed}{\xmark}  & {1.013} & \makecell{ViT-B/16 \\ \citep{dosovitskiy2020image}} & 87M & 11.2 \\ \cmidrule{2-6}
& \textcolor{SJViolet}{\cmark} & 1.001 & 
\makecell{ViT-B/16 \\ \citep{dosovitskiy2020image}} 
& 0.9M & 8.4 \\ \bottomrule
\end{tabular}%
}
\end{table}

In Table \ref{appendix-training-cost}, we compare the time complexity of different methods. Firstly, our \ALgname (without LoRA \citep{hulora}) largely outperforms Multi-scale Denoised \citep{jeong2024multi}, 
\ie 0.743 $\rightarrow$ 1.013 in ACR, with significantly smaller training costs, \ie  32 $\rightarrow$ 11.2 in GPU days. Furthermore, LoRA reduces the training time of \ALgname by 25\% and the trainable parameters by 99\% compared to full parameter fine-tuning, while maintaining ACR on par with full parameter fine-tuning. 
We note that the efficiency of LoRA can be further improved through advancements of GPU infrastructure or low-level code optimization. Since LoRA is generally applicable to other robust training techniques, we hope that our work initiates the research direction on alleviating the large cost of robust training.

Meanwhile, Diffusion Denoised \citep{carlinicertified} proposes to obtain a robust classifier \emph{without} fine-tuning. While they achieve reasonable robustness in an extremely efficient manner, \ie no fine-tuning costs, they suffer from the fundamental limitation associated with hallucination effect of the denoiser (see Figure~\ref{fig:hallucinated_images}). Due to this bottleneck, we find that the robustness of Diffusion Denoised degrades particularly at large radii (see Table \ref{imagenet-table}). One of our main contributions is identifying and addressing such hallucination issue,  achieving improved robustness, \ie 0.896 $\rightarrow$ 1.013 in ACR.

\clearpage
\section{Analysis on Pre-trained Classifiers }

\begin{table}[htb!]
\centering
\caption{Comparison of ACR and certified accuracy between \ALgname and Diffusion Denoised \citep{carlinicertified} on CIFAR-10 with $\sigma = 0.50$.}
\label{appendix-different-pretrained}
\begin{tabular}{cc}
    \centering
    \resizebox{0.9\textwidth}{!}{%
            \begin{tabular}{lcccccccccc}
            \toprule
                \multirow{2.5}{*}{{Method}} & 
                \multirow{2.5}{*}{{Classifier}} &
                \multirow{2.5}{*}{\shortstack{Test \\ Accuracy}} &
                \multirow{2.5}{*}{{ACR}} 
                & \multicolumn{7}{c}{Certified Accuracy at $\varepsilon$ (\%)} \\ \cmidrule{5-11}
                &&&& 0.00 & 0.25 & 0.50 & 0.75 & 1.00 & 1.25 & 1.50 \\
                \toprule
                \makecell[l]{Diffusion Denoised \\ \citep{carlinicertified}} & ResNet-110 (1.7M) & 93.7\% & 0.669 & \textbf{75.0} & 62.8 & 50.2 & 38.9 & 30.9 & 22.8 & 15.6 \\
                \midrule
                \multirow{2}{*}{\textbf{FT-CADIS (Ours)}} & ResNet-110 (1.7M) & 93.7\% & 0.754 & 68.8 & 60.9 & 53.9 & 45.2 & 38.3 & 29.4 & 23.4\\
                & ViT-B/16 (85.8M) & \textbf{97.9\%} & \textbf{0.806} & 72.2 & \textbf{64.1} & \textbf{57.2} & \textbf{48.1} & \textbf{40.3} & \textbf{34.1} & \textbf{25.9} \\
                
            \bottomrule
            \end{tabular}
        }
\end{tabular}
\end{table}

In Table \ref{appendix-different-pretrained}, we investigate the relationship between our proposed method and pre-trained classifiers. The results show that our proposed method still outperforms \citet{carlinicertified} on ResNet-110 \citep{he2016deep}, \ie a much smaller architecture than ViT-B/16 \citep{dosovitskiy2020image}. Also, we observe that the certified robustness of our method improves as we use more advanced pre-trained classifiers, \eg \ALgname based on ViT-B/16 largely improves the results based on ResNet-110. These findings demonstrate that the effectiveness of our method is not restricted to specific classifiers and can be further enhanced with continuous advancements in this field.

\section{Additional Results on \texorpdfstring{$\ell_\infty$}{Lg} Adversarial Attack}

\begin{table}[htb!]
\centering
\caption{Comparison of $\ell_\infty$ certified accuracy (\%) on CIFAR-10 with radius $\varepsilon$. We report the model with the highest certified $\ell_\infty$ accuracy for each method.}
\label{appendix-linf-table}
\resizebox{0.9\textwidth}{!}{%
\begin{tabular}{l|cc|c}
\toprule

CIFAR-10 ($\ell_\infty$) & \makecell{Diffusion Denoised \\ \citep{carlinicertified}} 
& \makecell{Multi-scale Denoised \\ \citep{jeong2024multi}}
& \makecell{\textbf{\ALgname (Ours)}} \\ 
\midrule
Robust  ($\varepsilon = \frac{2}{255}$)
& 62.9
& 67.1     
& \textbf{71.8} \\ 

\bottomrule
\end{tabular}
}
\end{table}

In Table \ref{appendix-linf-table}, we present the certified robustness of defense methods on another threat model, \ie $\ell_\infty$-norm. Specifically, we leverage the geometric relationships between the $\ell_2$-norm ball and $\ell_\infty$-norm ball to assess the robustness under $\ell_\infty$-norm \citep{salman2019provably}. Through this simple conversion, our proposed method can provide robustness against other $\ell_p$-adversaries.

\clearpage
\section{Additional Examples of Hallucinated Images}
\begin{figure}[htb!]
    \centering
    \begin{subcaptionblock}{0.81\textwidth}
        \centering        \includegraphics[width=0.81\linewidth]{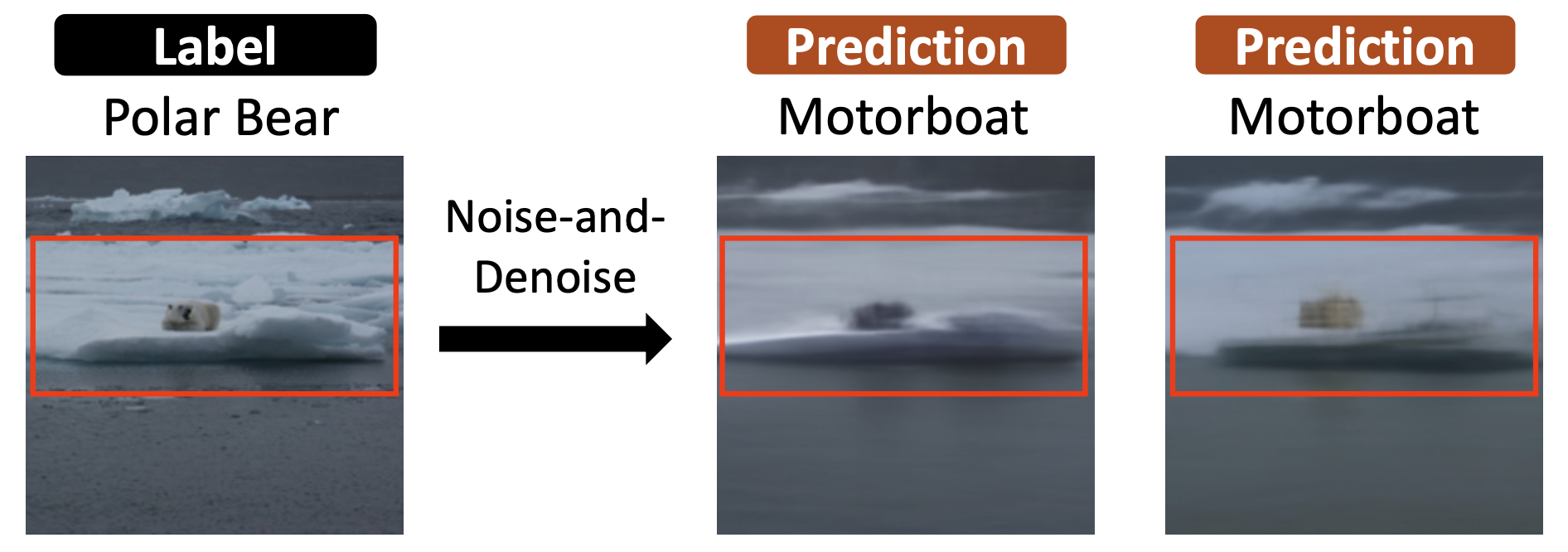}
        \label{fig:hallucinated_polarbear}
    \end{subcaptionblock}
    
    \vspace{0.2cm}
    
    \begin{subcaptionblock}{0.81\textwidth}
        \centering        \includegraphics[width=0.81\linewidth]{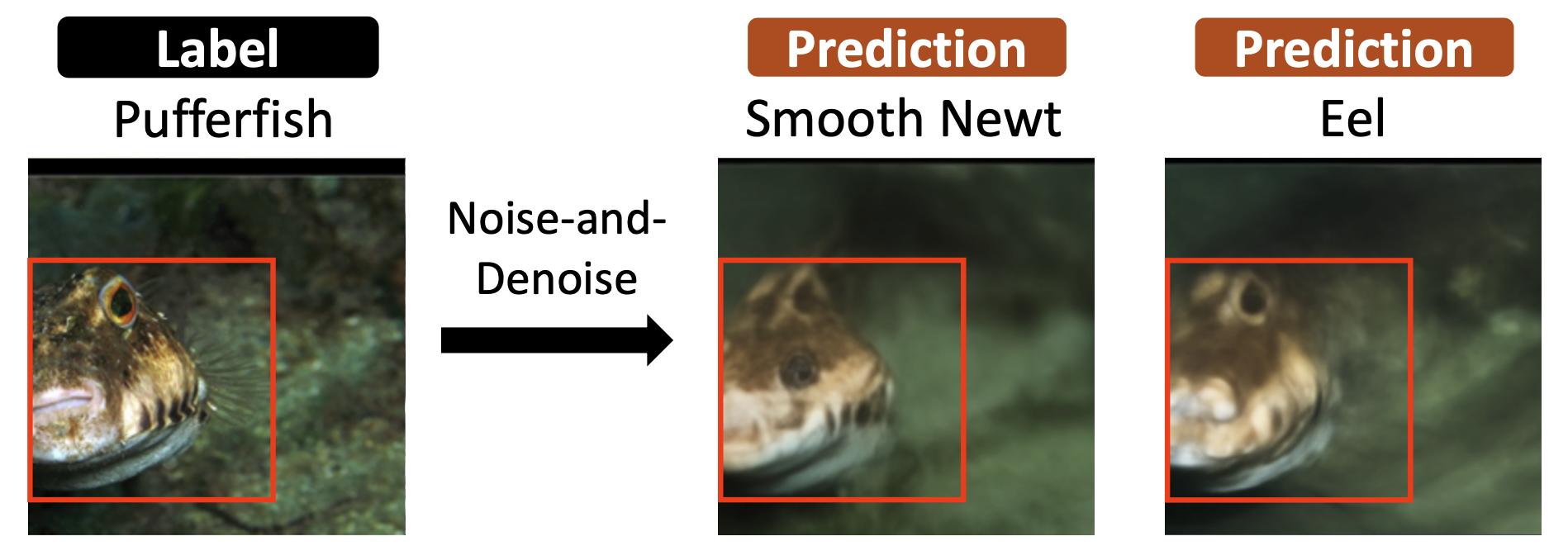}
        \label{fig:hallucinated_pufferfish}
    \end{subcaptionblock}
    
    \vspace{0.2cm}
    
    \begin{subcaptionblock}{0.81\textwidth}
        \centering        \includegraphics[width=0.81\linewidth]{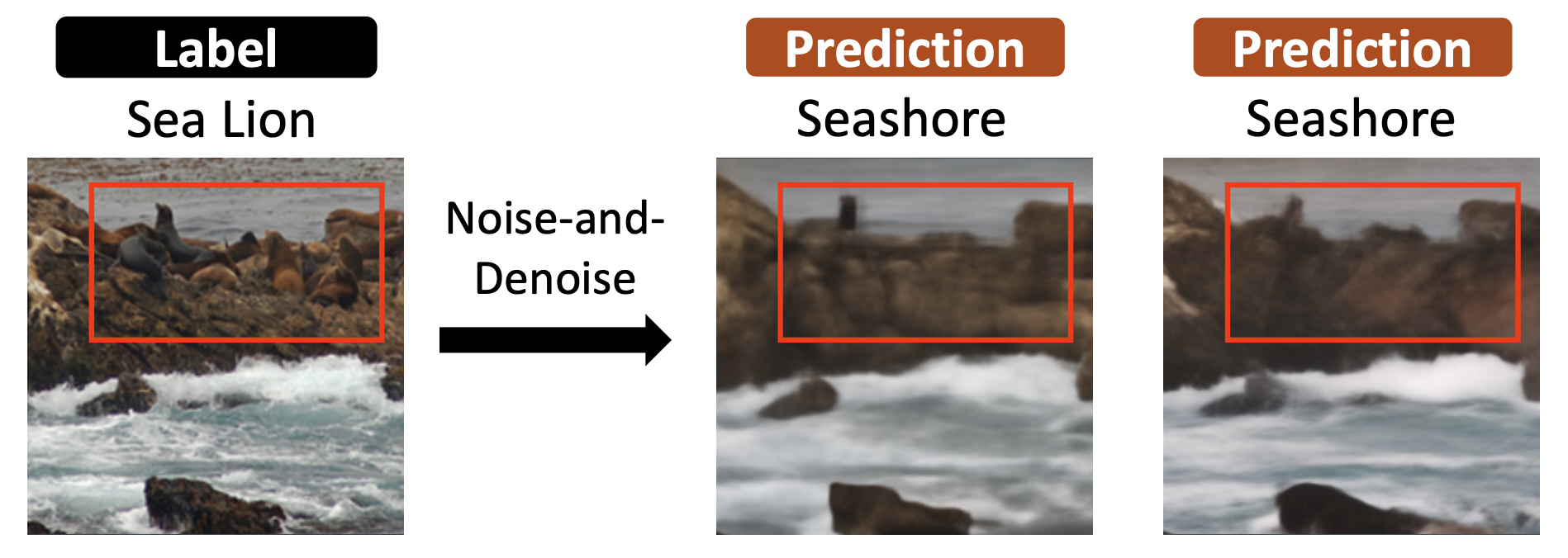}
    \label{fig:non_hallucinated_sealion}
    \end{subcaptionblock}
    
    \vspace{0.2cm}
    
    \begin{subcaptionblock}{0.81\textwidth}
        \centering        \includegraphics[width=0.81\linewidth]{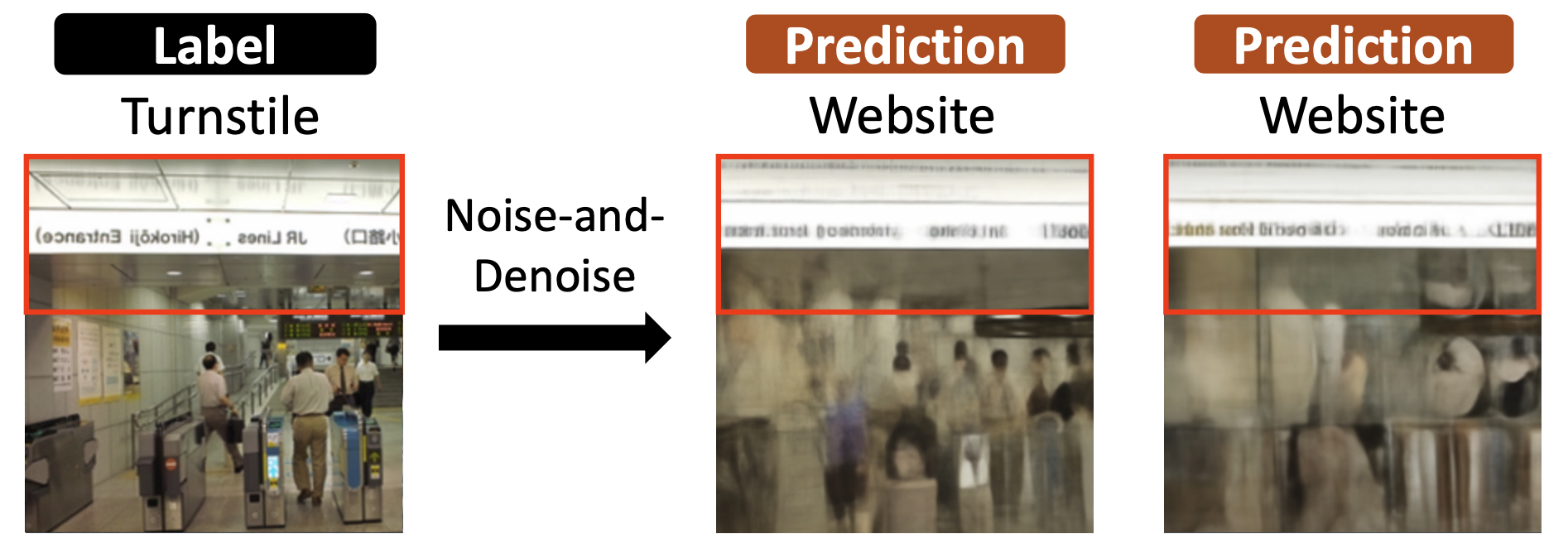}
\label{fig:non_hallucinated_turnstile}
    \end{subcaptionblock}
    \vspace{0.2cm}
    
    \begin{subcaptionblock}{0.81\textwidth}
        \centering        \includegraphics[width=0.81\linewidth]{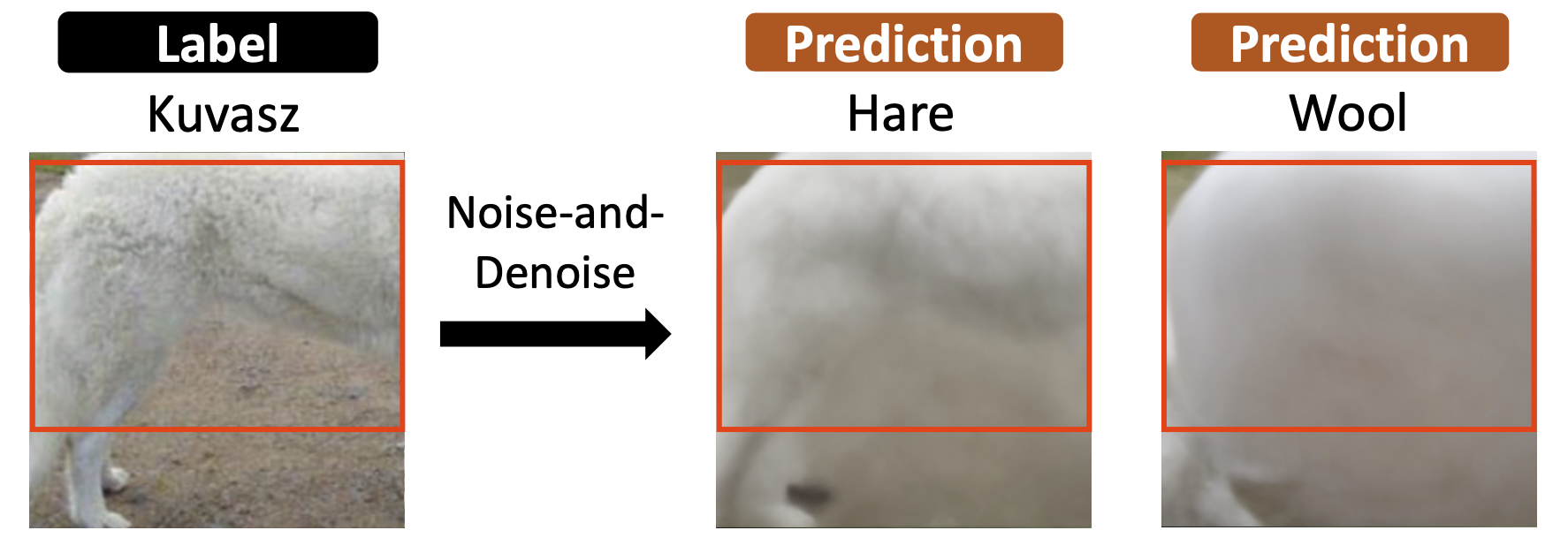}
\label{fig:non_hallucinated_kuvasz}
    \end{subcaptionblock}

    \caption{Additional examples of hallucinated images after the noise-and-denoise procedure on ImageNet at $\sigma=1.00$. The red box indicates the areas where the original semantic of the image is corrupted.}
    \label{fig:hallucinated_images_imagenet}
\end{figure}

\begin{figure}[htb!]
    \centering
    \begin{subcaptionblock}{0.81\textwidth}
        \centering        \includegraphics[width=0.81\linewidth]{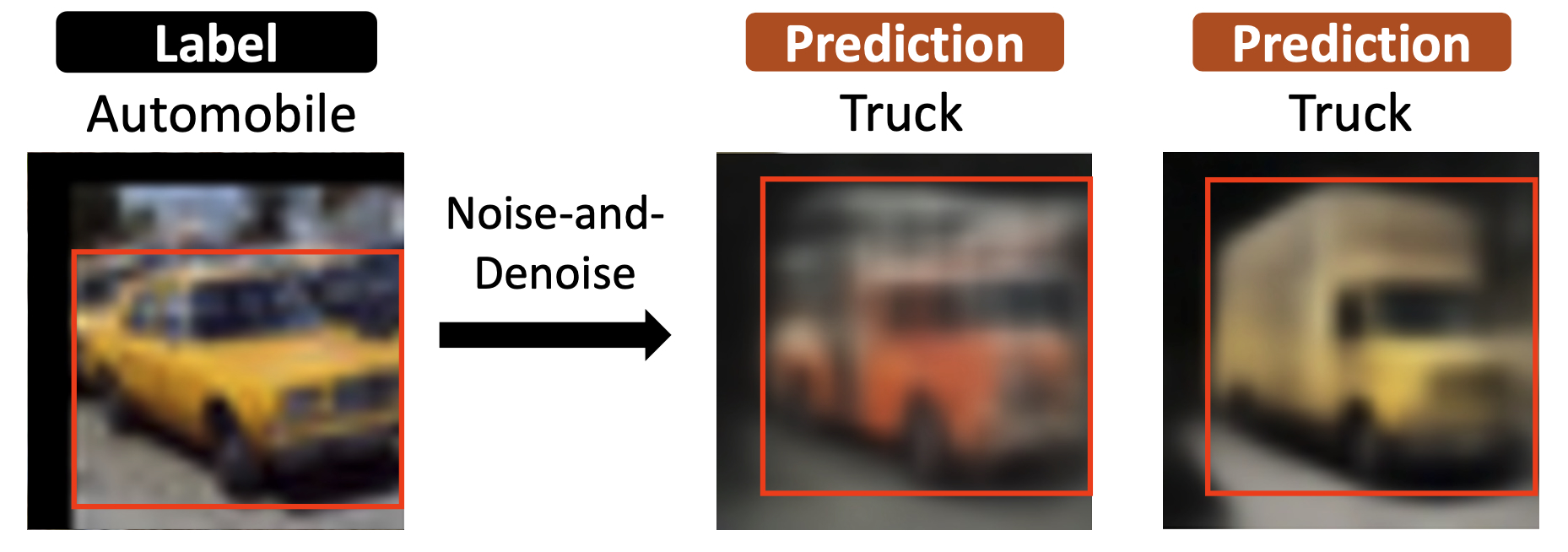}
        \label{fig:hallucinated_automobile}
    \end{subcaptionblock}
    
    \vspace{0.2cm}
    
    \begin{subcaptionblock}{0.81\textwidth}
        \centering        \includegraphics[width=0.81\linewidth]{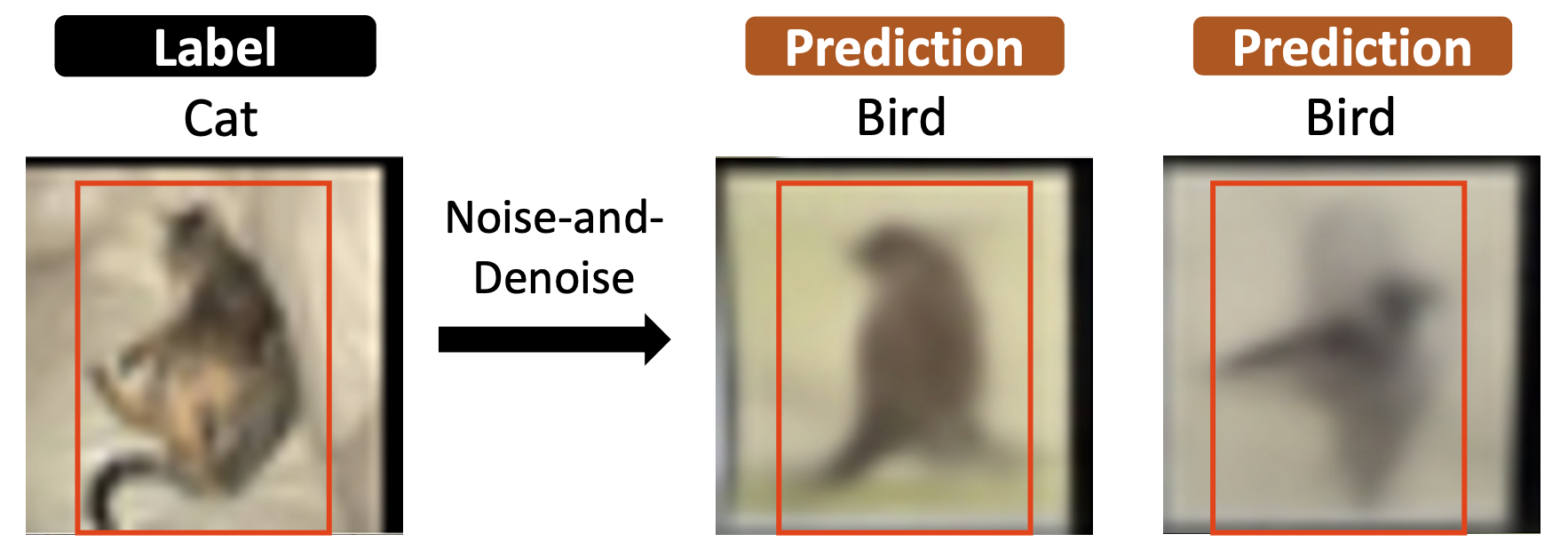}
        \label{fig:hallucinated_cat}
    \end{subcaptionblock}
    
    \vspace{0.2cm}
    
    \begin{subcaptionblock}{0.81\textwidth}
        \centering        \includegraphics[width=0.81\linewidth]{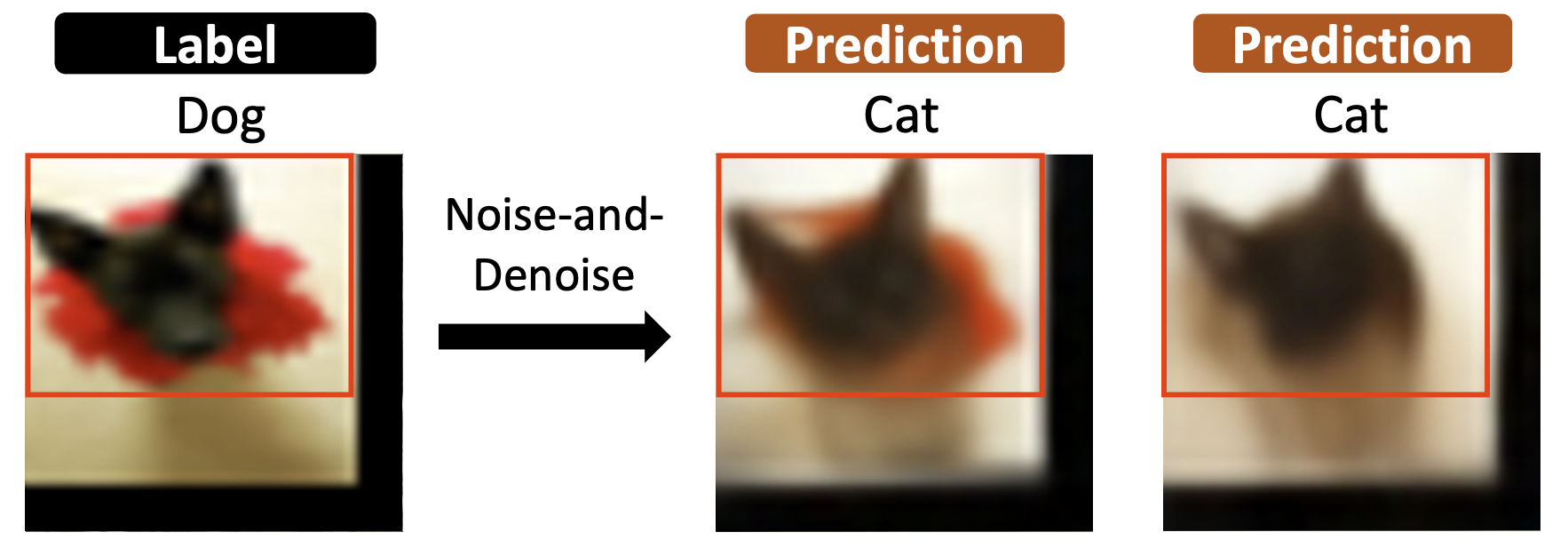}
    \label{fig:non_hallucinated_dog}
    \end{subcaptionblock}
    
    \vspace{0.2cm}
    
    \begin{subcaptionblock}{0.81\textwidth}
        \centering        \includegraphics[width=0.81\linewidth]{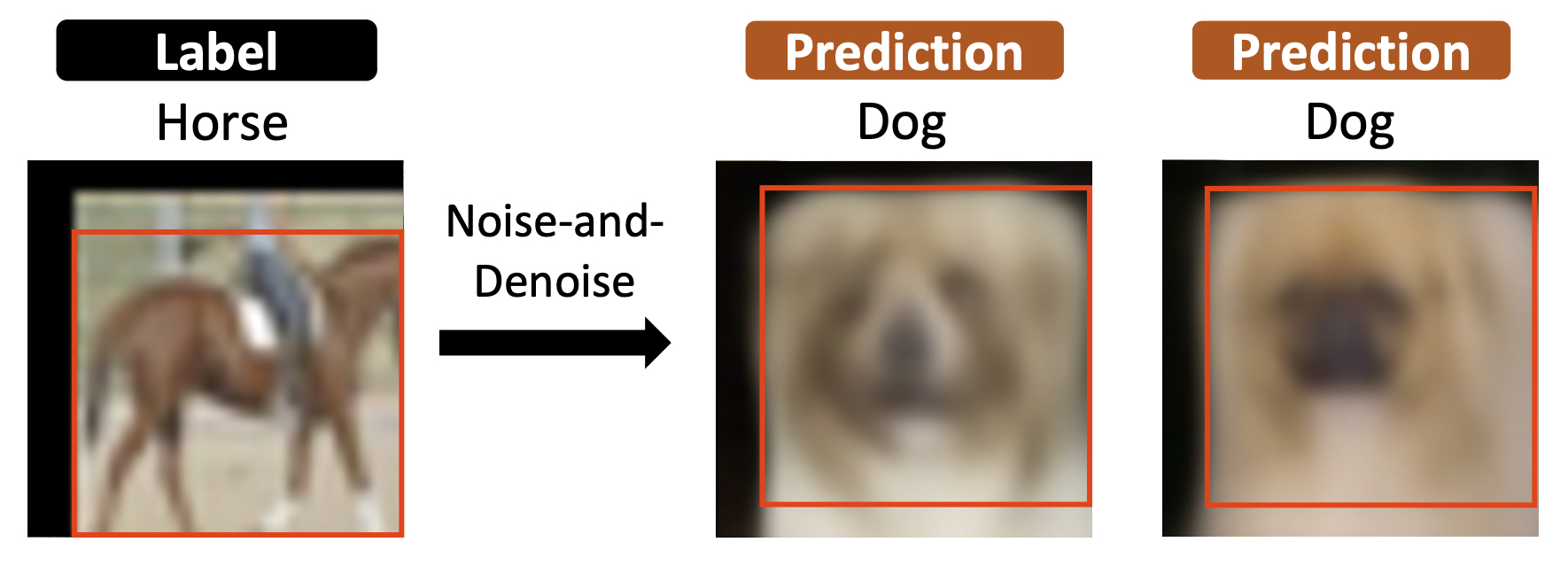}
\label{fig:non_hallucinated_horse}
    \end{subcaptionblock}
    \vspace{0.2cm}
    
    \begin{subcaptionblock}{0.81\textwidth}
        \centering        \includegraphics[width=0.81\linewidth]{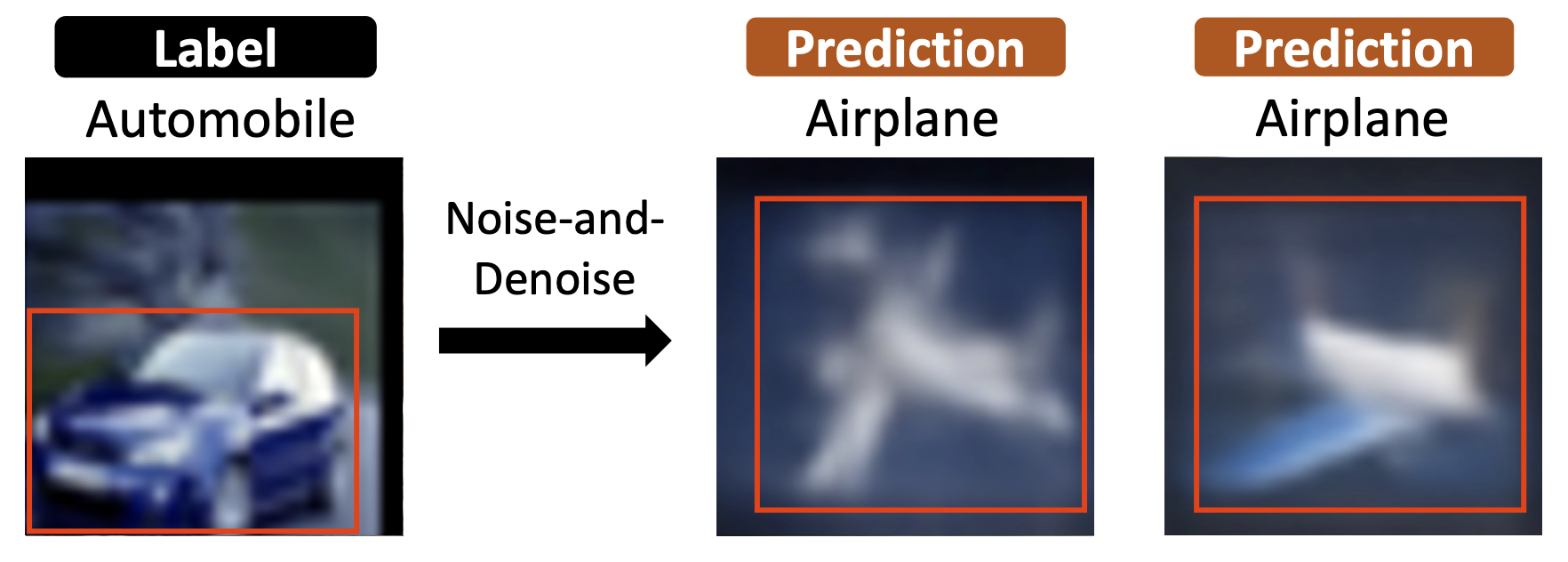}
\label{fig:non_hallucinated_automobile2}
    \end{subcaptionblock}

    \caption{Additional examples of hallucinated images after the noise-and-denoise procedure on CIFAR-10 at $\sigma=1.00$. The red box indicates the areas where the original semantic of the image is corrupted.}
    \label{fig:hallucinated_images_cifar10}
\end{figure}

\clearpage
\section{Analysis on Training Stability of FT-CADIS}

\begin{figure}[htb!]
    \centering
    \begin{minipage}[b]{0.45\textwidth}
        \centering
        \includegraphics[width=\textwidth]{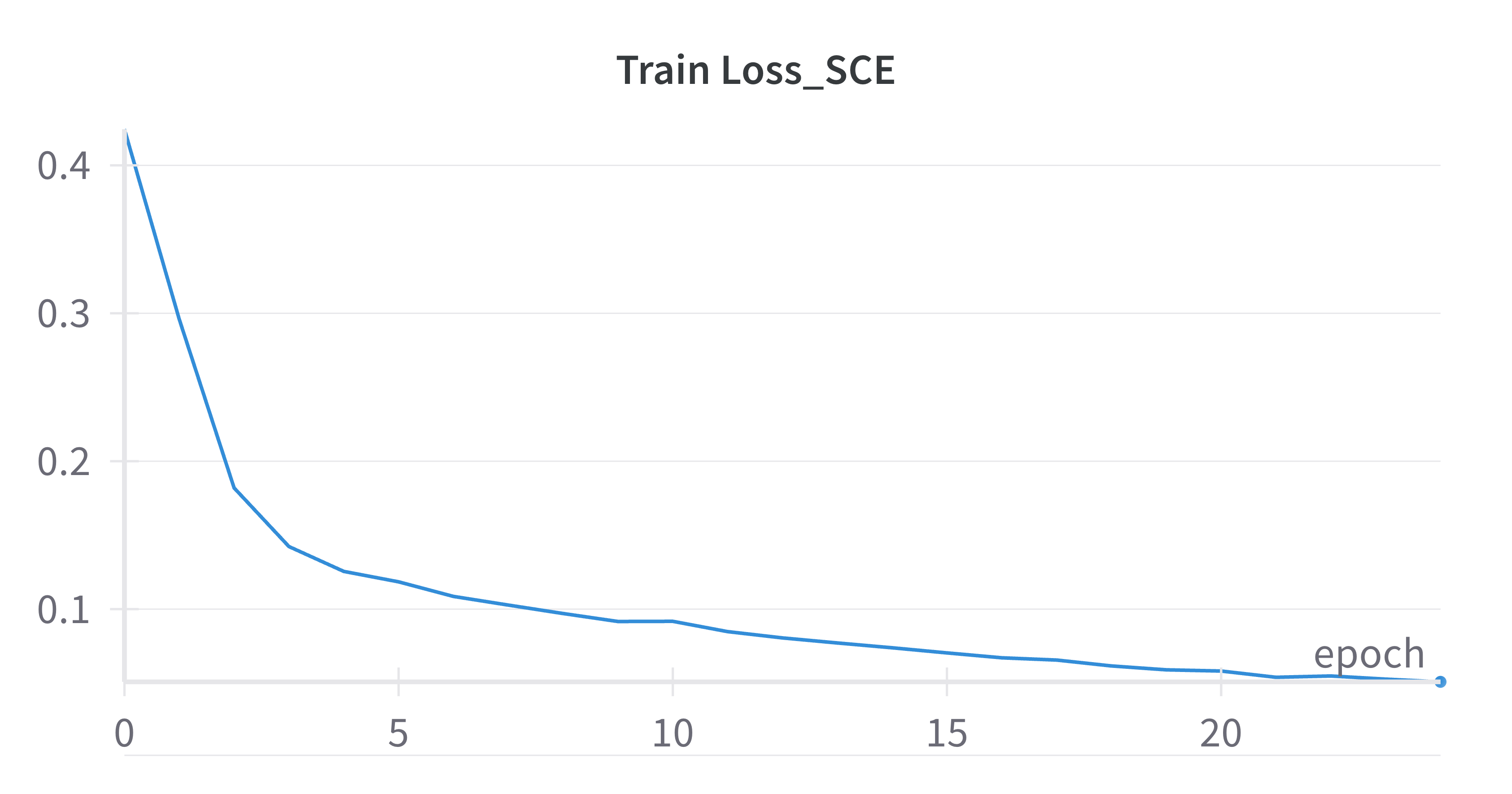}
    \end{minipage}
    \hfill
    \begin{minipage}[b]{0.45\textwidth}
        \centering
        \includegraphics[width=\textwidth]{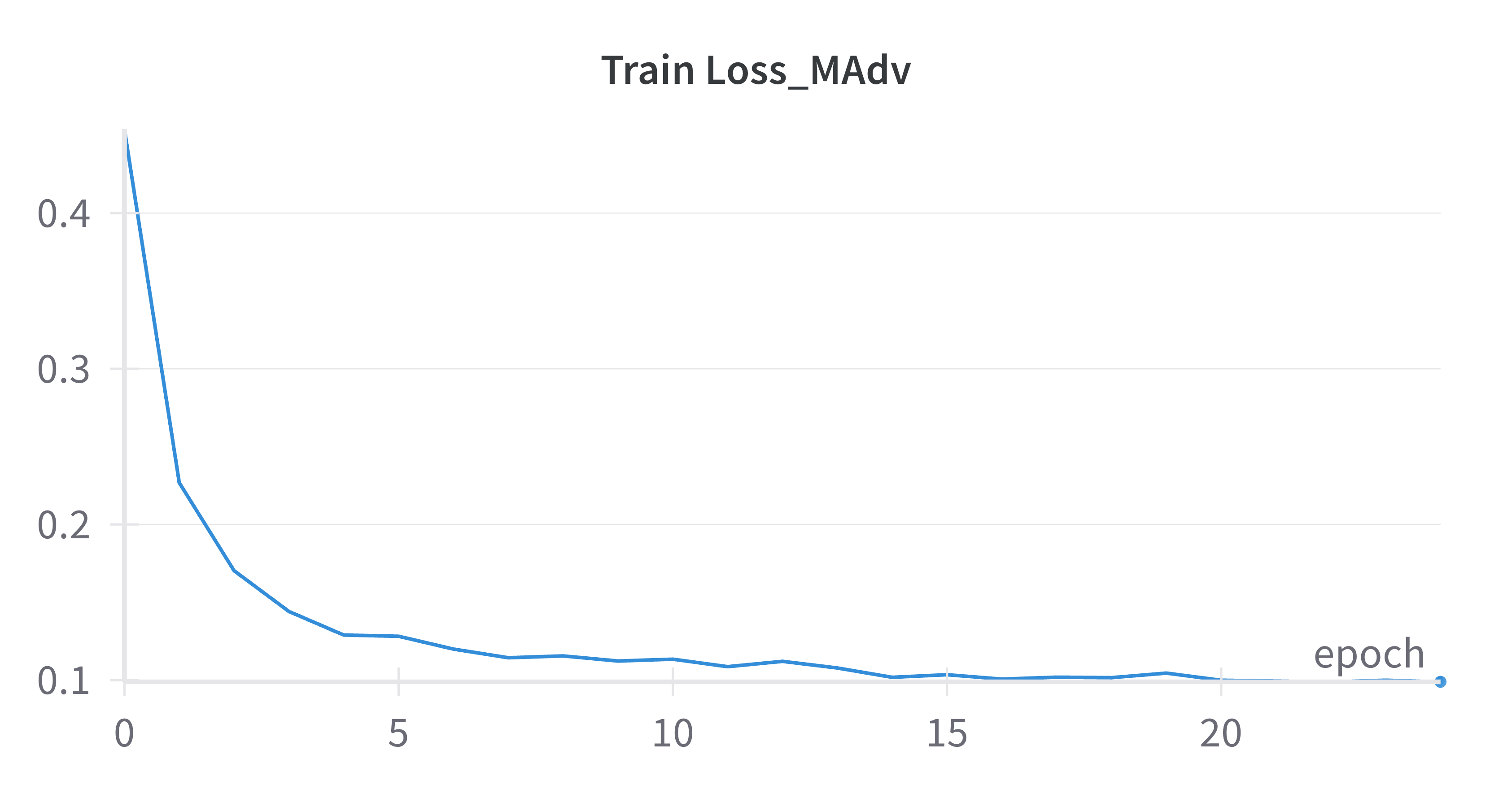}
    \end{minipage}
    
    \vspace{0.5cm} 
    
    \begin{minipage}[b]{0.45\textwidth}
        \centering
        \includegraphics[width=\textwidth]{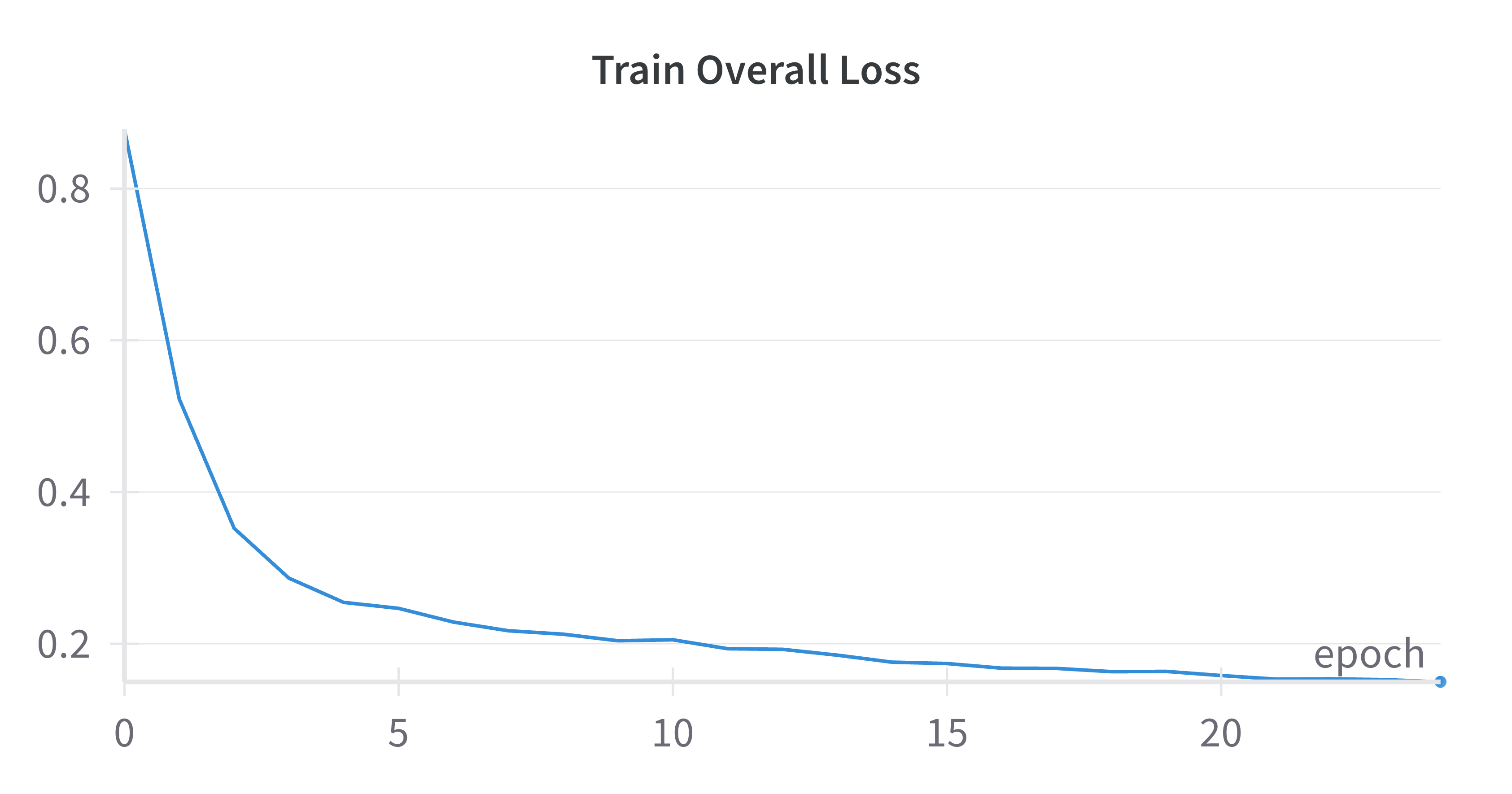}
    \end{minipage}
    \hfill
    \begin{minipage}[b]{0.45\textwidth}
        \centering
        \includegraphics[width=\textwidth]{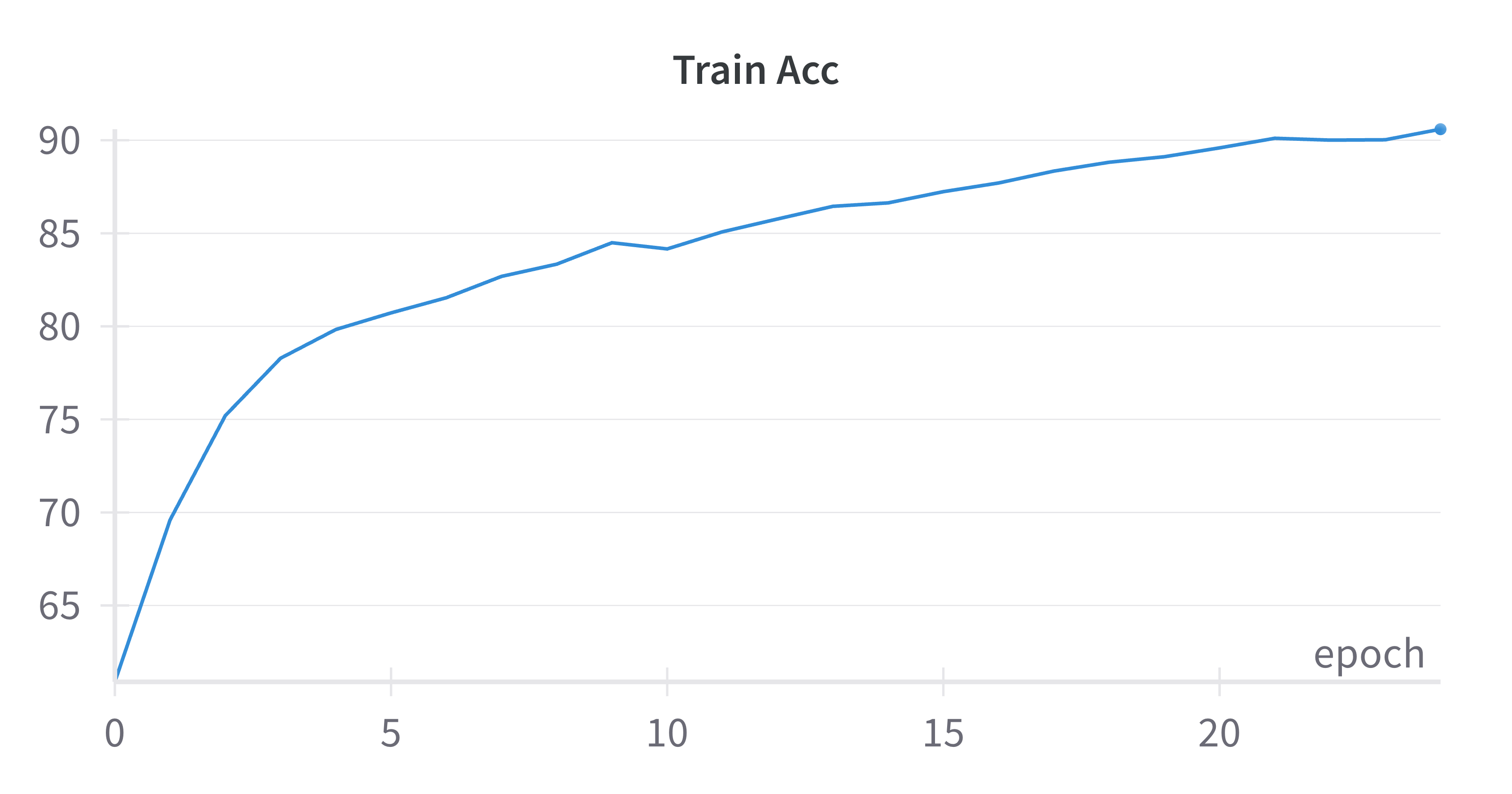}
    \end{minipage}
    
    \caption{Plots of (1) \LowLoss $\mathcal{L}^{\mathtt{SCE}}$, (2) \HighLoss $\mathcal{L}^{\mathtt{MAdv}}$, (3) Overall training objective $\mathcal{L}^{\mathtt{\ALgname}}$, and (4) Top-1 accuracy from our main experiments on CIFAR-10 with $\sigma = 0.25$.}
    \label{figure:loss_graph}
\end{figure}

In this section, we demonstrate that our training objective $\mathcal{L}^{\mathtt{\ALgname}}$ remains stable throughout the fine-tuning process. As mentioned in Section \ref{sec:FT-CADIS}, our overall objective is composed of \LowLoss $\mathcal{L}^{\mathtt{SCE}}$ and \HighLoss $\mathcal{L}^{\mathtt{MAdv}}$. In Figure \ref{figure:loss_graph}, we show that (1) the training loss, including the adversarial loss, converges smoothly without oscillation, and (2) the training accuracy also converges well.

\end{document}